\documentclass[sigconf,nonacm]{acmart}

\usepackage{amsmath}
\usepackage{mathtools}
\usepackage{graphicx}
\usepackage{subcaption}
\usepackage{booktabs}
\usepackage{multirow}
\usepackage{array}
\usepackage{tabularx}
\usepackage{algorithm}
\usepackage{algpseudocode}
\usepackage{siunitx}
\usepackage{booktabs}
\usepackage{longtable}
\usepackage{xcolor}
\usepackage{xspace}
\usepackage{enumitem}

\newcommand{\proj}{\textit{SolarChain-Eval}\xspace}

\AtBeginDocument{%
  }

\makeatletter
\renewcommand{\@fnsymbol}[1]{%
  \ensuremath{%
    \ifcase#1\or \dagger\or *\or \ddagger\or \mathsection\or \mathparagraph\or \|\or **\or \dagger\dagger\or \ddagger\ddagger \else\@ctrerr\fi%
  }%
}
\makeatother

\begin{document}

\title{\proj: A Physics-Constrained Benchmark for Trustworthy Economic Agents in Decentralized Energy Markets}

\newcommand{\sharedaffiliation}{%
  \affiliation{%
    \institution{Duke Kunshan University}
    \city{Kunshan}
    \state{Jiangsu}
    \country{China}
  }%
}

\author{Shilin Ou}
\sharedaffiliation

\authornotemark[1]
\author{Yifan Xu}
\sharedaffiliation
\authornote{Equal contribution.}

\author{Luyao Zhang}
\sharedaffiliation

\authornote{The Corresponding author: Email: lz183@duke.edu, Digital Innovation Research Center and Social Science Division, Duke Kunshan University, Address: Duke Avenue No.8, Kunshan, Suzhou, Jiangsu, China, 215316. \textbf{Acknowledgments}: Shilin Ou is grateful for the support from the Summer Research Scholar Program at Duke Kunshan University, supervised by Prof. Luyao Zhang.}

\renewcommand{\shortauthors}{Ou et al.}

\begin{abstract}
As agentic AI systems are increasingly applied to cyber-physical environments, their evaluation requires assessment of both task performance and trustworthiness. In decentralized energy markets, autonomous agents may improve market utility, but may also exploit invalid physical data, create artificial liquidity, and produce unstable governance decisions. Therefore, we propose \proj, a physics-constrained benchmark for evaluating trustworthy economic agents. It formulates market governance as a Gymnasium-compatible Markov Decision Process, where agents make hourly decisions. \proj evaluates each policy across multiple dimensions, including market utility, physical safety, slippage, action smoothness, spatial fairness, and auditability.

To support agentic evaluation, \proj incorporates an LLM-based Planner/Auditor layer. The Planner defines episode-level action bounds and audit rules, while the Auditor reviews and revises high-risk actions. All interventions are recorded through structured logs, including trigger signals, proposed actions, revised actions, and audit rationales. Experiments with static, random, myopic, RL, and RL+LLM policies reveal a clear utility-safety trade-off. RL agents improve market utility but can still produce unsafe behavior. When the physics penalty is removed, reward-maximizing agents exploit invalid generation and increase artificial liquidity. The LLM Planner/Auditor improves auditability and mitigates selected risks, but it cannot fully compensate for a misspecified reward function. These results indicate that trustworthy agentic AI evaluation requires both physical constraints and transparent intervention traces. We release data and code as open access on GitHub for replicability.
\end{abstract}

%

\begin{teaserfigure}
    \centering
    \includegraphics[width=\textwidth]{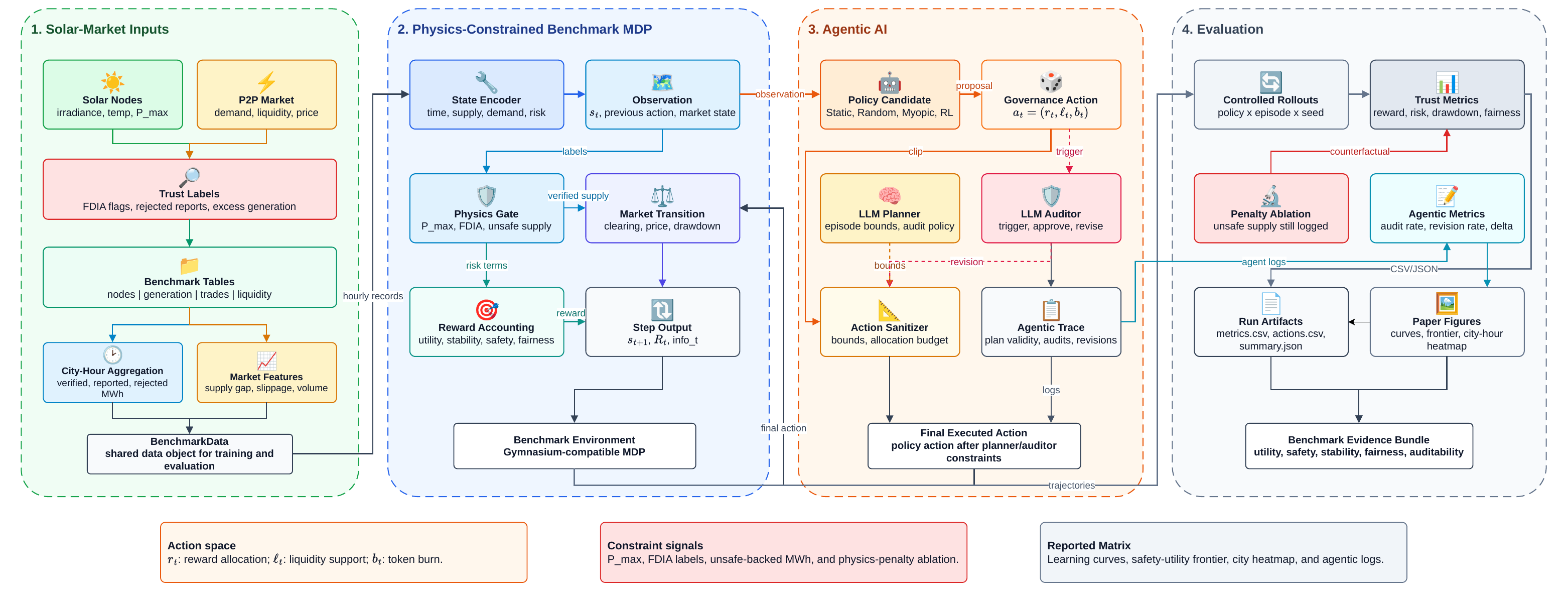}
    \caption{Overall \proj benchmark pipeline. The figure summarizes the end-to-end workflow from data loading,  RL training, policy evaluation, output generation, and trustworthiness evidence for decentralized energy-market governance.}
    \label{fig:overall_pipeline}
\end{teaserfigure}

\maketitle

\section{Introduction}
\label{sec:intro}

Agentic AI is increasingly used to support autonomous decision-making in cyber-physical and economic systems\cite{xi2025rise}. Such agents not only optimize digital objectives, their actions may also affect physical resources, market incentives, and systems safety\cite{vyas2025autonomous}. This shift raises an evaluation challenge that agent performance cannot be judged by scalar reward alone. Trustworthy agents should therefore make decisions that are physically grounded, constraint-aware, and auditable under real-world operating conditions \cite{nist2023airmf,drgona2025safe,phiri2025auditable}.

\begin{table*}[t]
\centering
\caption{Comparison of Prevailing Bi-Level Architectures Integrating RL and LLM Agents}
\renewcommand{\arraystretch}{1.3}
\begin{tabular*}{1\textwidth}{@{\extracolsep{\fill}} p{2.5cm} p{2.5cm} p{5.5cm} p{4.5cm} @{}}
\toprule
\textbf{Framework} & \textbf{Domain} & \textbf{Synergy Mechanism (LLM + RL)} & \textbf{Limitations \& Challenges} \\
\midrule
\textbf{RE-GoT}~\cite{yao2025reward} & Robotic Control \& Manipulation & Upper-level LLM uses Graph-of-Thoughts to evolve reward functions; Vision models analyze RL rollouts for semantic feedback. & High latency due to LLM/VLM inference bottlenecks in real-time control loops.\\
\addlinespace
\textbf{$\text{RL}^2$}~\cite{yang2025rl2} & Active Distribution Networks & LLM translates operator safety intents into mathematical penalties; adapts parameters via DRL feedback. & Sensitive to LLM hallucination during  algebraic adjustments. \\
\addlinespace
\textbf{ADMM-LLM}~\cite{yang2025llm} & Microgrid Heat-Electricity & LLM acts as an optimizer that observes optimization residuals to predict and adjust decentralized ADMM penalty parameters. & Heuristic trial-and-error lacks strict mathematical convergence guarantees. \\
\bottomrule
\end{tabular*}
\label{tab:agent_rl_comparison}
\end{table*}

In decentralized energy markets, reinforcement learning (RL) has been used to manage market liquidity, token issuance, and incentive distribution \cite{tushar2018transforming,ou2026solarchain}. However, unlike purely digital environments, energy markets are bounded by physical laws, such as localized solar irradiance and photovoltaic panel capacities \cite{zhang2018review,mengelkamp2018designing}. Deploying reward-maximizing agents in such environments therefore creates important trustworthiness risks. Without explicit physical constraints, RL models can learn to exploit weaknesses in the market mechanism \cite{achiam2017constrained}. 

To address this, we propose \proj, a physics-constrained benchmark for evaluating trustworthy agentic AI in decentralizedenergy markets. \proj combines solar  signals with controlled market and risk scenarios, and formulates market governance as a Gymnasium-compatible Markov Decision Process. To further examine agentic oversight, \proj incorporates an evaluation-time LLM-based Planner/Auditor layer, which is inserted between the trained RL policy and the benchmark environment without being used during RL training. The Planner defines episode-level action bounds and audit rules, while the Auditor reviews and revises high-risk actions. This design enables \proj to assess both the operational performance of an RL policy, and the extent to which its decision-making behavior remains governable and interpretable.

Specifically, we concentrate on the following research questions:

\begin{itemize}
\item \textbf{RQ1:} How do autonomous market policies trade off economic utility, physical safety, stability, and fairness compared with static and heuristic baselines?

\item \textbf{RQ2:} To what extent do reward-maximizing agents exploit invalid generation and create artificial liquidity when physics penalties are removed?

\item \textbf{RQ3:} How does an LLM-based Planner/Auditor contribute to auditable risk mitigation for trained RL policies, and what are its limitations under reward misspecification?
\end{itemize}

\noindent\textbf{Data and Code Availabilty Statements}: We release data and code as open access on GitHub\footnote{\url{https://github.com/yxu-dev/SolarChain-Eval}} for replicability.
\section{Literature Review}

\subsection{Reinforcement Learning in Tokenomics}

Traditional economic mechanisms struggle when applied to decentralized energy markets due to the restrictive assumptions of stationary environments~\cite{zheng2022ai}.  Distributed Energy Resources (DERs) are governed by intermittent generation constraints and Peer-to-Peer (P2P) network topologies~\cite{chen2022peer}. However, tokenomic mechanisms are subject to extreme volatility driven by liquidity depth, speculative token price trajectories, and adoption rates~\cite{malinova2023tokenomics, cong2021tokenomics}. RL provides a flexible modeling approach by formulating market governance as a Markov Decision Process~\cite{zheng2022ai}. It can adapt to high-dimensional state spaces, optimizing long-term objectives that balance physical utility with economic stability\cite{zheng2022ai}.

In current decentralized market governance literature\cite{xu2025auto,qu2025rules}, macroeconomic control is typically abstracted into continuous governance levers, such as the reward ratio ($r_t$), liquidity ratio ($l_t$), and token burn rate ($b_t$). In simulated environments, the normalized action vector $a_t$ is decoded into specific economic parameters. To evaluate system trustworthiness, researchers widely rely on standardized frameworks like \textit{Stable-Baselines3}~\cite{raffin2021stable} to benchmark distinct RL algorithmic paradigms:

Proximal Policy Optimization (PPO) is an on-policy actor-critic algorithm widely applied to continuous action space governance tasks. To prevent excessively large policy updates, PPO utilizes a clipped surrogate objective~\cite{schulman2017proximal}:
    \begin{equation}
        L^{CLIP}(\theta) = \hat{\mathbb{E}}_t \left[ \min(p_t(\theta)\hat{A}_t, \text{clip}(p_t(\theta), 1-\epsilon, 1+\epsilon)\hat{A}_t) \right]
    \end{equation}

\textit{where $\theta$ denotes the policy network parameters, $p_t(\theta)$ is the probability ratio, $\hat{A}_t$ represents the estimated advantage function, and $\epsilon$ is the clipping hyperparameter.}

In tokenomic governance, this high training stability translates to practical advantages by effectively suppressing action jitter and ensuring smoother adjustments of macroeconomic parameters.
    
Also supporting continuous control, Soft Actor-Critic (SAC) is an off-policy algorithm. By incorporating a maximum entropy mechanism into its objective function~\cite{haarnoja2018soft}, SAC proactively encourages the agent to explore diverse tokenomic strategies:
    \begin{equation}
        J(\pi) = \sum_{t=0}^{T} \mathbb{E}_{(s_t, a_t) \sim \rho_\pi} \left[ r(s_t, a_t) + \alpha \mathcal{H}(\pi(\cdot|s_t)) \right]
    \end{equation}

\textit{where $r(s_t, a_t)$ denotes the immediate reward, $\alpha$ is the temperature parameter, and $\mathcal{H}(\pi(\cdot|s_t))$ represents the entropy of the policy $\pi$ at state $s_t$. }
    
Existing research demonstrates that this mechanism not only improves sample efficiency but also prevents the policy from prematurely converging to suboptimal local minima in highly nonlinear market environments\cite{campos2022soft}.
    
As a robust value-based baseline algorithm, Deep Q-Network (DQN) is applied to discrete action spaces. In multidimensional control tasks, continuous levers are typically discretized. Despite inherent quantization errors, DQN excels at accurately estimating the action-value function $Q(s, a)$~\cite{mnih2015human,van2016deep}.

\begin{figure*}[t]
    \centering
    \includegraphics[width=\textwidth]{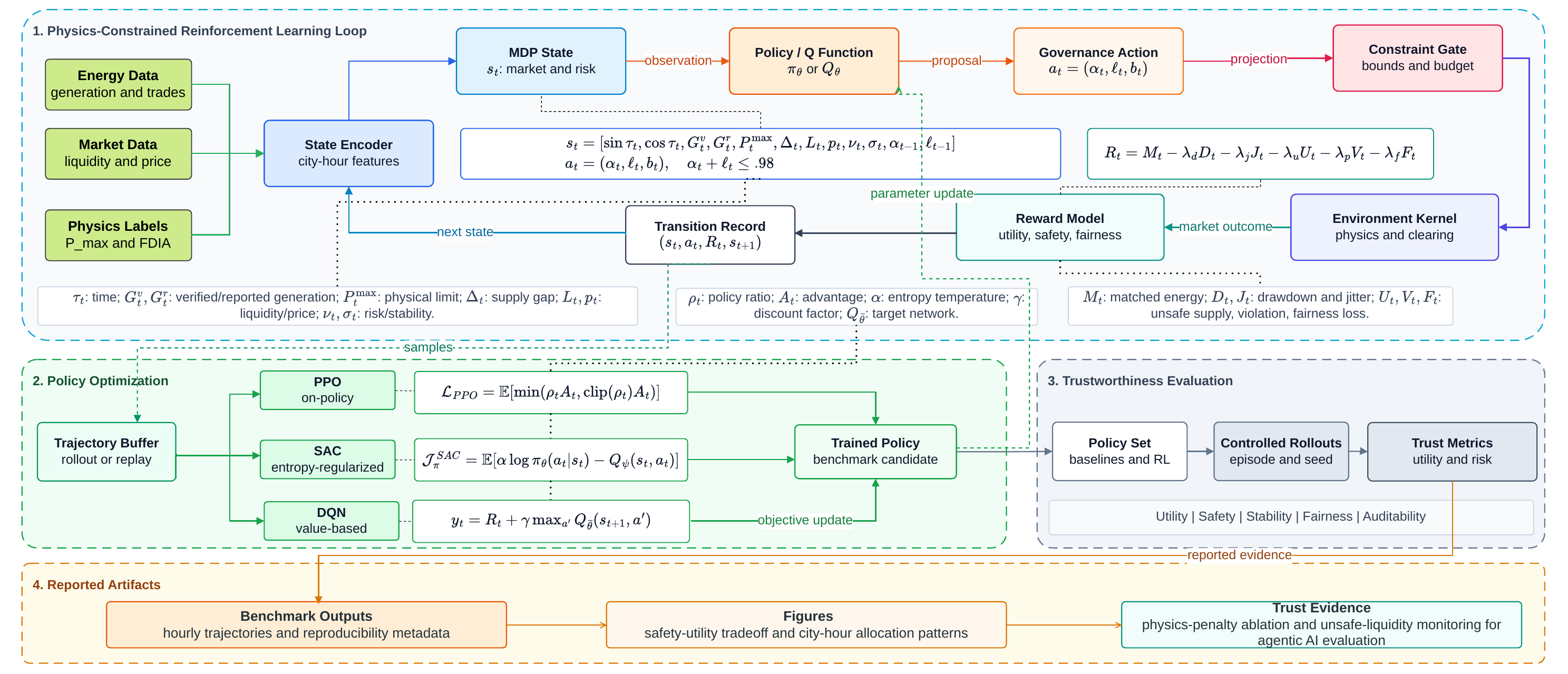}
    \caption{Reinforcement learning evaluation framework for SolarChain-Eval. The diagram connects market and physical inputs to policy actions, environment feedback, PPO/SAC/DQN updates, and trustworthiness-oriented evaluation across utility, safety, stability, smoothness, and fairness.}
    \label{fig:rl_method_framework}
\end{figure*}

\subsection{RL \& LLM Agent Integration}

RL has shown strong performance in high-dimensional continuous control tasks. However, its black-box nature limits its reliability in infrastructure~\cite{amiri2025deep}. In practice, human auditors may find it difficult to explain why an RL policy makes adjustments~\cite{rudin2019stop}. Furthermore, while post-hoc  AI  methods offer retrospective insights, they cannot proactively prevent unsafe actions\cite{gu2024review}.

Recent studies address this by integrating Large Language Models (LLMs) with RL systems. LLMs could act as high-level reasoners to process task context and operator instructions~\cite{yao2022react}. This ability allows them to translate operator intents into mathematical penalties that can support lower-level decision making~\cite{du2023guiding}.

Consequently, current research is shifting toward a bi-level RL-LLM architecture~\cite{carta2023grounding}. The lower layer uses RL policies to make frequent control decisions based on value estimates such as $V(s)$ or $Q(s,a)$. The upper layer uses LLMs for slower and more abstract reasoning tasks. Prior work has used LLMs to convert operator intent into reward penalties~\cite{kwon2023reward}, revise reward functions, and tune optimization parameters~\cite{ma2023eureka}. Table~\ref{tab:agent_rl_comparison} summarizes representative examples of this design.

Recent advances in RL have improved agent-based optimization in Cyber-Physical Systems, especially in decentralized energy markets~\cite{kelly2020rl,lozano2025democratizing}. As the focus shifts toward real-world deployment~\cite{dulac2021challenges}, there is a growing need to complement traditional reward-centric benchmarks with evaluations of trustworthiness. Specifically, ensuring safe operation requires assessing an agent's adherence to physical constraints and its resilience in adversarial scenarios~\cite{liu2011false}. Following this view, our benchmark couples economic dynamics with physical ground truths and evaluates agentic trustworthiness across multiple dimensions.

\section{Methodology}
\label{sec:methodology}

Instead of merely maximizing short-term trading volume, our benchmark evaluates the overall trustworthiness of an autonomous economic agent. It assesses whether the agent can maintain market utility while respecting physical limits, ensuring market stability, reducing action jitter, and preserving spatial fairness.
Each episode represents a 24 hour market cycle, sampled from a 720-hour dataset from April 2026. Episodes are constrained to start at the beginning of the day. This temporal design ensures a consistent daily decision horizon for the agent while preserving real-world weather patterns.

\subsection{RL Formulation}

Figure~\ref{fig:rl_method_framework} summarizes the RL loop. Energy generation, market demand, and physics labels are aggregated into an hourly benchmark state. Based on this state, a policy proposes a governance action. 

The state $s_t$ and action $a_t$ are defined as:
\begin{equation}
\label{eq:state-action}
\begin{aligned}
s_t &= [\sin\tau_t, \cos\tau_t, G_t^v, G_t^r, P_t^{\max}, \Delta_t, L_t, p_t, \nu_t, \sigma_t, \alpha_{t-1}, \ell_{t-1}], \\
a_t &= (\alpha_t, \ell_t, b_t), \qquad \alpha_t + \ell_t \leq 0.98.
\end{aligned}
\end{equation}
\textit{Here, $\tau_t$ is the time angle, $G_t^v$ is verified generation, $G_t^r$ is reported generation, $P_t^{\max}$ is the physical PV upper bound, $\Delta_t$ is the supply-demand gap, $L_t$ is market liquidity, $p_t$ is token price, $\nu_t$ is the raw risk signal, and $\sigma_t$ is the static slippage. }

The action $a_t$ comprises the reward allocation ratio $\alpha_t$, liquidity injection ratio $\ell_t$, and token burn rate $b_t$. Standard algorithms like PPO and SAC operate on the continuous action space, while DQN utilizes a discretized $5^3=125$ action grid over the same parameters. All trained policies are evaluated under the uniform benchmark setting detailed in Table~\ref{tab:benchmark_setting}.

\begin{table}[htbp]
\centering
\small
\caption{Main benchmark configuration.}
\label{tab:benchmark_setting}
\begin{tabular}{l l}
\toprule
\textbf{Item} & \textbf{Value} \\
\midrule
Cities & Beijing, Shanghai, Chengdu, Shenzhen, Hangzhou \\
Energy nodes & 50 \\
Main period & 2026-04-01 to 2026-04-30 \\
Hourly market states & 720 \\
Generation records & 36,000 \\
P2P trade records & 1,185 \\
Episode length & 24 hours \\
Evaluated Policies & Static, Random, Myopic, PPO, SAC, DQN \\
\bottomrule
\end{tabular}
\end{table}

\subsection{Physics Limit and Market Transition}

\proj implements a physics constraint module, which verifies whether reported generation is physically feasible and FDIA labels. Let $X_t$ denote suspicious or invalid supply. The benchmark not only records this anomaly but quantifies the extent to which the policy economically backs it:
\begin{equation}
\label{eq:physics-gate}
V_t = \frac{\beta_t X_t}{\max(G_t^v + \beta_t X_t, \epsilon)}, \qquad \beta_t = \frac{\min(\alpha_t + \ell_t, 0.98)}{0.98}.
\end{equation}
In Eq.~\eqref{eq:physics-gate}, $V_t$ is the action-dependent physics violation term. A policy incurs higher risk when it heavily backs invalid supply. This formulation directly addresses RQ2: through a no-physics-penalty ablation, we can observe whether a purely reward-driven agent exploits invalid generation to inflate artificial liquidity.

Additionally, the market module processes the backed supply, consisting of verified supply and the action-backed suspicious supply. It updates the matched energy $M_t$ and computes the residual liquidity $L_{t+1}$ for the next hour:
\begin{equation}
\label{eq:market-clearing}
\begin{aligned}
M_t &= \min(L_t + \ell_t(G_t^v + \beta_t X_t),\ \widetilde{Q}_t), \\
L_{t+1} &= \max(L_t + \ell_t(G_t^v + \beta_t X_t) - M_t, 0).
\end{aligned}
\end{equation}
Here, $\widetilde{Q}_t$ represents the effective demand after token burning, computed as
\[
\widetilde{Q}_t = Q_t^d \max(0.75, 1 - 0.80b_t),
\]
where \(Q_t^d\) is the raw market demand. This mechanism explicitly links the agent's governance decisions to both trading volume and market liquidity depth.

\subsection{Reward Design and RL Optimization}

To optimize for both utility and market trustworthiness, the step reward implements a series of penalties (Figure~\ref{fig:rl_method_framework}):
\begin{equation}
\label{eq:reward}
\begin{aligned}
R_t &= M_t - \lambda_d D_t - \lambda_j J_t - \lambda_u U_t
      - \lambda_p \Phi_t - \lambda_f F_t, \\
\Phi_t &= V_t + A_t, \qquad A_t=\beta_t X_t.
\end{aligned}
\end{equation}
The positive term $M_t$ rewards successful market clearing. The corresponding penalties address liquidity drawdown ($D_t$), action jitter ($J_t$), unmet demand ($U_t$), composite physics risk ($\Phi_t$), and spatial unfairness across cities ($F_t$). Here, \(\Phi_t\) combines the normalized physics violation rate \(V_t\) and the action-backed suspicious supply amount \(A_t\).

The RL policies are trained using Stable-Baselines3. The benchmark incorporates three learned controllers with distinct optimization biases: PPO for stable on-policy updates, SAC for entropy-guided continuous exploration, and DQN for discrete Q-value learning. For instance, DQN optimizes the following target over the discretized action grid:
\begin{equation}
\label{eq:dqn-target}
y_t = R_t + \gamma \max_{a'} Q_{\bar{\theta}}(s_{t+1}, a').
\end{equation}
Static, random, and myopic controllers are also evaluated as non-learning baselines to provide a comparison.

\subsection{LLM Planner/Auditor Layer}
\label{sec:agent_governance}

\begin{figure*}[t]
    \centering
    \includegraphics[width=\textwidth]{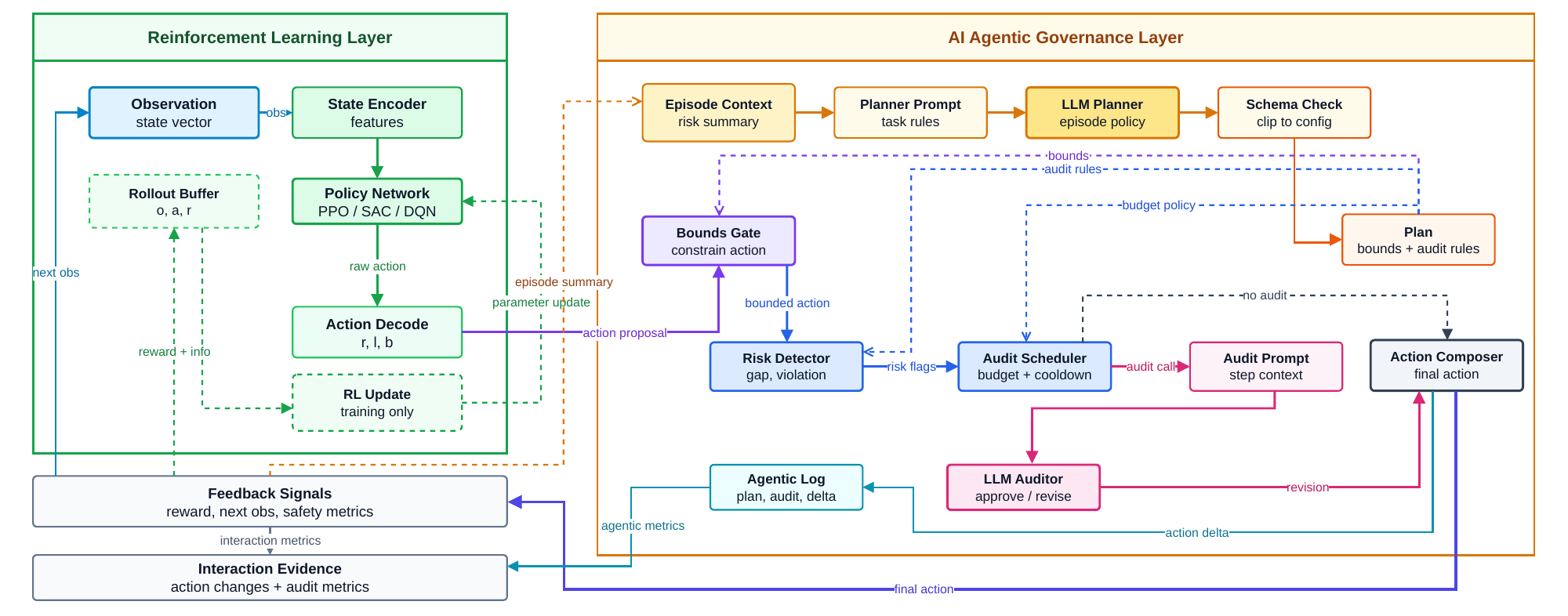}
    \caption{Interaction between the reinforcement learning layer and the AI-agent governance layer. The RL policy proposes market actions, which are bounded by the Planner and conditionally revised by the Auditor under high-risk scenarios.}
    \label{fig:rl_agent_interaction}
\end{figure*}

To evaluate whether agentic oversight improves deployment-time safety, we introduce an AI-agent governance layer powered by LLMs, as shown in Figure~\ref{fig:rl_agent_interaction}. We insert this layer between the trained RL policy (e.g., PPO, SAC, DQN) and the market environment. This layer is active only during evaluation and is never used during RL training. This separation ensures that any observed safety improvement stems from deployment-time guardrails rather than from reward shaping, policy retraining, or parameter updates.

The governance layer consists of two components: a Planner and an Auditor.

The Planner operates at the episode level. At the start of each 24-hour episode, it reviews episode-level context, including supply-demand adequacy, physics-risk statistics, static-market slippage, and benchmark action constraints. Based on this context, the Planner establishes admissible bounds for the RL action components, namely the reward ratio \(\alpha_t\), liquidity ratio \(\ell_t\), and burn rate \(b_t\). It also outputs an episode-specific audit policy, including risk thresholds, a maximum audit budget \(B_e\), a target audit rate, and a minimum cooldown length \(c_e\). At each hourly step, the RL policy's proposed action is first decoded into market variables and clipped by the Planner-defined bounds and benchmark constraints, yielding a bounded action \(\bar{a}_t\).

The Auditor provides sparse step-level oversight. Instead of reviewing every action, it is invoked only when risk signals indicate possible safety or market-integrity failures. To avoid notational conflict with the reward terms in Eq.~\eqref{eq:reward}, we denote the audit-side action-instability signal by \(\kappa_t\). It is computed as the \(\ell_1\) distance between the current bounded action and the previous executed action:
\begin{equation}
\label{eq:agentic-jitter}
\kappa_t = \|\bar{a}_t - a_{t-1}\|_1 .
\end{equation}
We further define the audit-side normalized supply-demand gap as
\begin{equation}
\label{eq:supply-demand-gap}
\Delta^{audit}_t = \frac{G_t^v-Q_t^d}{\max(Q_t^d,\epsilon)},
\end{equation}
where \(G_t^v\) denotes verified energy supply, \(Q_t^d\) denotes market demand/load, and \(\epsilon>0\) avoids division by zero. Negative values of \(\Delta^{audit}_t\) indicate supply shortfall.

Audit triggers are divided into hard and soft categories. Hard triggers respond to immediate safety or market-integrity threats, including excessive physical violation rate \(\nu_t\), supply shortfall \(\Delta^{audit}_t\), and high static-market slippage \(\sigma_t\):

\begin{equation}
\label{eq:hard-trigger}
H_t =
\mathbb{I}\left[
\nu_t > \tau_{\nu}
\ \vee\
\Delta^{audit}_t < \tau_{\Delta}
\ \vee\
\sigma_t > \tau_{\sigma}
\right].
\end{equation}
Soft triggers capture erratic controller behavior through action instability:
\begin{equation}
\label{eq:soft-trigger}
S_t =
\mathbb{I}\left[
\kappa_t > \tau_{\kappa}
\right].
\end{equation}

An action becomes audit-eligible when either trigger fires:

\begin{equation}
\label{eq:event-trigger}
E_t = H_t \vee S_t .
\end{equation}

To prevent excessive LLM intervention, soft-triggered audits are rate-limited by the episode budget and cooldown. In contrast, hard triggers bypass these limits due to their severity. Let \(N^{audit}_{e,t}\) denote the number of audits already used in episode \(e\) before step \(t\), and let \(t^{last}_e\) denote the most recent audited step. The final audit decision is

\begin{equation}
\label{eq:audit-trigger}
\mathbb{I}^{audit}_t
=
E_t
\wedge
\left[
H_t
\ \vee\
\left(
N^{audit}_{e,t} < B_e
\ \wedge\
t - t^{last}_e > c_e
\right)
\right].
\end{equation}

Thus, severe physics or market failures can trigger immediate review, while softer instability signals are subject to sparse intervention limits.

If \(\mathbb{I}^{audit}_t=0\), the market directly executes the bounded action \(\bar{a}_t\). If \(\mathbb{I}^{audit}_t=1\), the Auditor receives the current observation, the bounded proposed action, the previous action, the Planner policy, and recent market diagnostics. It then returns a decision: \texttt{approve}, which keeps \(\bar{a}_t\), or \texttt{revise}, which replaces it with a conservative corrected action. The wrapper deterministically applies this decision before calling the environment step.

Finally, to make the governance layer auditable and economically valid, all LLM outputs must follow  structured schemas and numerical validation(Appendix \ref{app:raw-data-provenance}: Table \ref{tab:llm-governance-protocol}). Planner outputs are clipped to valid benchmark action ranges and audit-policy ranges. Auditor outputs are sanitized against market constraints, including individual bounds on \(\alpha_t\), \(\ell_t\), and \(b_t\), as well as the global allocation constraint \(\alpha_t+\ell_t\leq 0.98\). This schema-enforced interface prevents free-form text from directly controlling market execution and ensures that every agentic intervention remains traceable and reproducible during evaluation.

\subsection{Evaluation Metrics}

\proj evaluates each policy from two perspectives: market utility and system trustworthiness. We compare six baselines: Static, Random, Myopic, PPO, SAC, and DQN. Static uses fixed governance parameters, Random samples actions uniformly, and Myopic follows a greedy rule based on current supply, demand, and risk signals. PPO and SAC operate in the continuous action space, while DQN uses the discretized action grid.

We use three evaluation settings:
\begin{itemize}
    \item \textbf{Main Benchmark:} evaluates all baselines using the full reward in Eq.~\eqref{eq:reward}. This setting tests whether RL policies can improve utility while maintaining safety, stability, smoothness, and fairness.
    \item \textbf{Reward Ablation:} removes the physics penalty from Eq.~\eqref{eq:reward}, while still logging physical violations. This setting tests whether agents exploit unsafe generation when physical risk is not penalized.
    \item \textbf{Agentic Evaluation:} adds the AI-agent governance layer to trained RL policies during evaluation. This setting tests whether LLM-based oversight can revise unsafe actions without retraining the RL controller.
\end{itemize}

\begin{table}[htbp]
\centering
\small
\caption{Trustworthiness-oriented evaluation dimensions and reported evidence.}
\label{tab:metric_dimensions}
\begin{tabular}{l l}
\toprule
\textbf{Dimension} & \textbf{Reported Evidence} \\
\midrule
Utility & Cumulative reward, episode trading volume \\
Physics Safety & Physics violation rate, artificial liquidity \\
Market Stability & Liquidity drawdown, token drawdown, slippage \\
Action Smoothness & Action jitter across hourly decisions \\
Spatial Fairness & City-level reward allocation imbalance \\
Auditability (Agentic) & Plan validity, audit rate, revision rate, action delta \\
\bottomrule
\end{tabular}
\end{table}
\begin{table*}[h]
\centering
\caption{Main benchmark performance under the physics-constrained reward. Values are mean $\pm$ standard deviation over all three seeds and 90 rollouts per policy. Higher cumulative reward and trading volume are better; lower physics violation, slippage, and artificial liquidity indicate safer market governance.}
\label{tab:main-performance}
\begin{tabular}{lccccc}
\toprule
Policy & Cumulative Reward & Trading Volume & Physics Violation Rate & Mean Slippage & Artificial Liquidity \\
\midrule
Static & -23.38 $\pm$ 0.80 & 0.514 $\pm$ 0.099 & 0.4874 $\pm$ 0.0166 & 0.0140 $\pm$ 0.0020 & 0.1803 $\pm$ 0.0419 \\
Random & -25.46 $\pm$ 0.91 & 0.480 $\pm$ 0.093 & 0.4771 $\pm$ 0.0172 & 0.0148 $\pm$ 0.0023 & 0.1137 $\pm$ 0.0297 \\
Myopic & -23.08 $\pm$ 0.80 & 0.456 $\pm$ 0.088 & 0.4779 $\pm$ 0.0168 & 0.0144 $\pm$ 0.0023 & 0.1009 $\pm$ 0.0235 \\
PPO & -22.35 $\pm$ 1.02 & 0.470 $\pm$ 0.097 & 0.4520 $\pm$ 0.0219 & 0.0134 $\pm$ 0.0023 & 0.1453 $\pm$ 0.0840 \\
SAC & -22.26 $\pm$ 0.94 & 0.521 $\pm$ 0.100 & 0.4461 $\pm$ 0.0196 & 0.0178 $\pm$ 0.0032 & 0.0686 $\pm$ 0.0328 \\
DQN & -23.23 $\pm$ 0.91 & 0.491 $\pm$ 0.096 & 0.4688 $\pm$ 0.0182 & 0.0148 $\pm$ 0.0023 & 0.1298 $\pm$ 0.0397 \\
\bottomrule
\end{tabular}
\end{table*}

\begin{figure*}[t]
\centering
\begin{subfigure}[t]{0.32\textwidth}
\centering
\includegraphics[width=\linewidth]{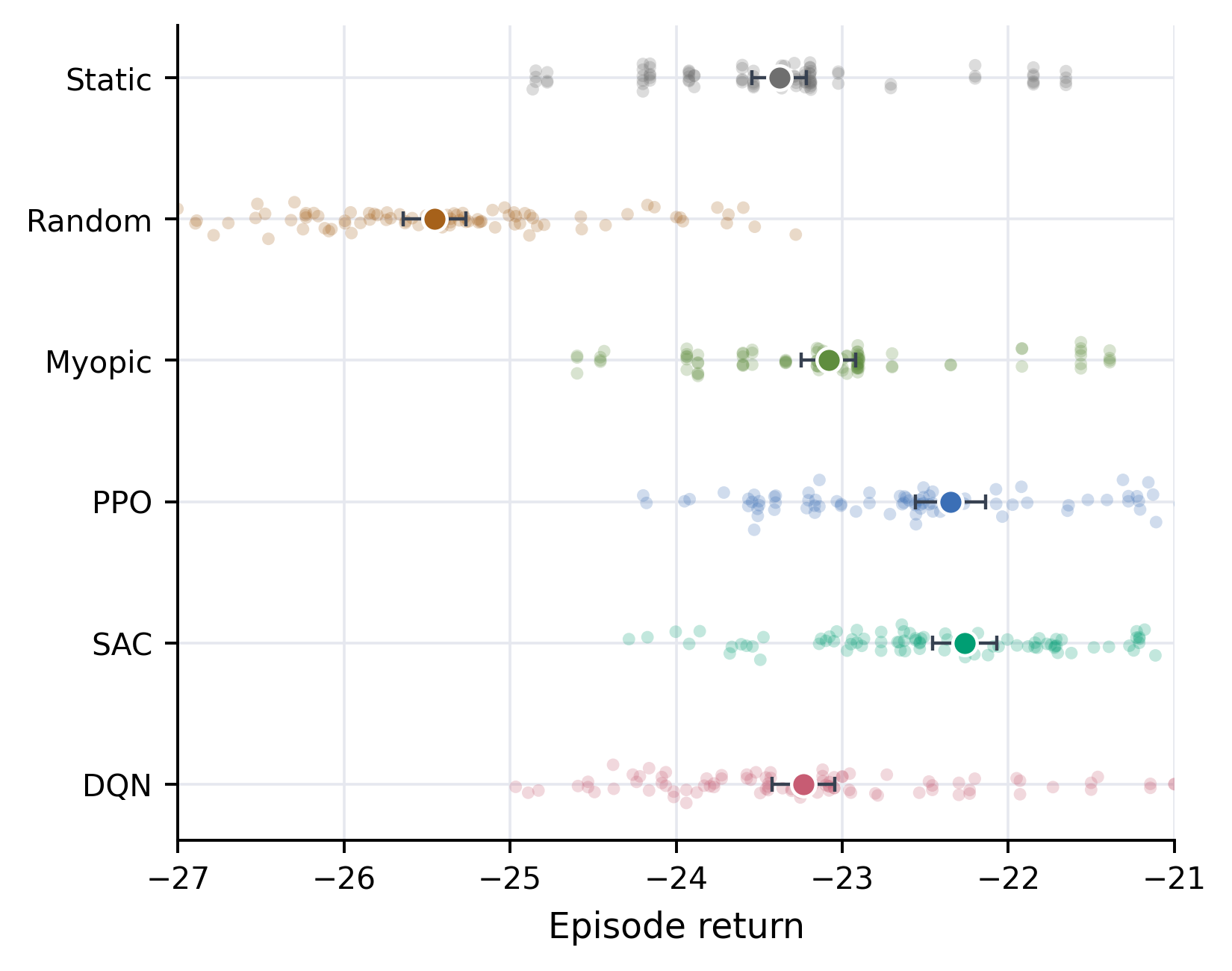}
\caption{Rollout-level cumulative reward.}
\label{fig:rq1-reward-distribution}
\end{subfigure}
\hfill
\begin{subfigure}[t]{0.32\textwidth}
\centering
\includegraphics[width=\linewidth]{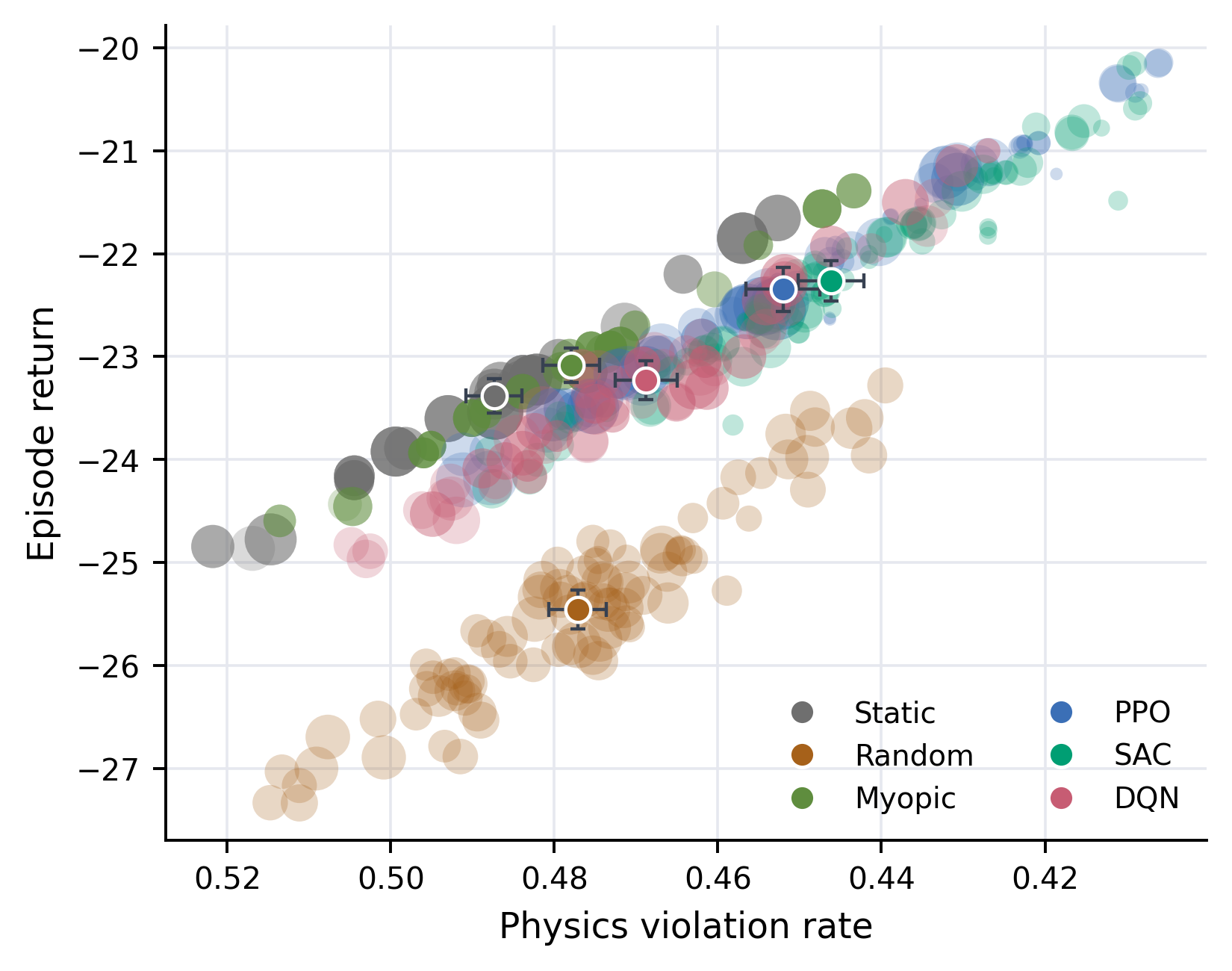}
\caption{Utility--safety frontier. Each point is one rollout; larger markers indicate higher artificial liquidity.}
\label{fig:rq1-utility-safety-frontier}
\end{subfigure}
\hfill
\begin{subfigure}[t]{0.32\textwidth}
\centering
\includegraphics[width=\linewidth]{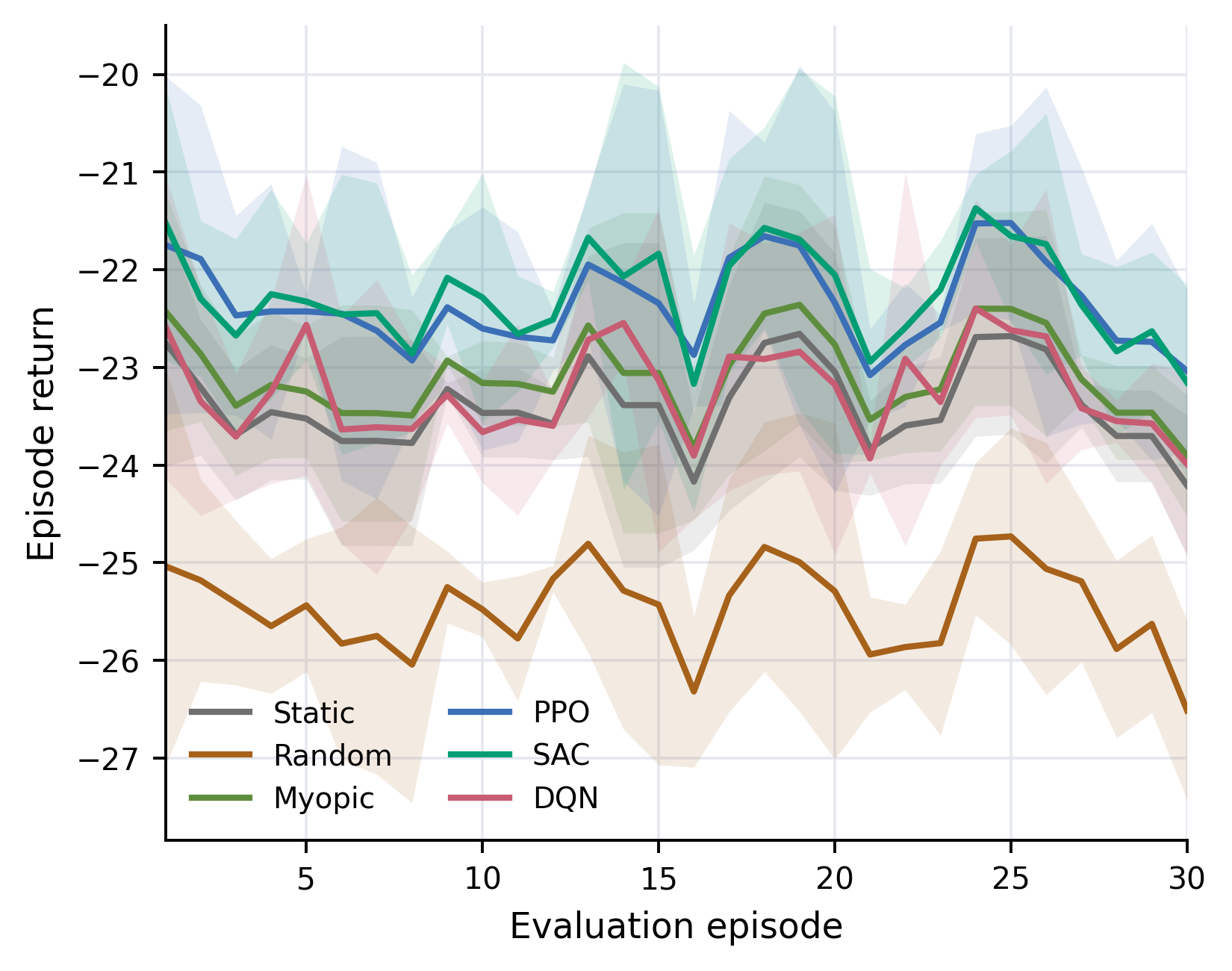}
\caption{Episode-wise reward trajectory with 95\% confidence bands across seeds.}
\label{fig:rq1-reward-trajectory}
\end{subfigure}
\caption{Main benchmark performance under the physics-constrained reward.}
\label{fig:rq1-main-benchmark}
\end{figure*}

\begin{figure*}[t]
\centering
\begin{subfigure}[t]{0.32\textwidth}
\centering
\includegraphics[width=\linewidth]{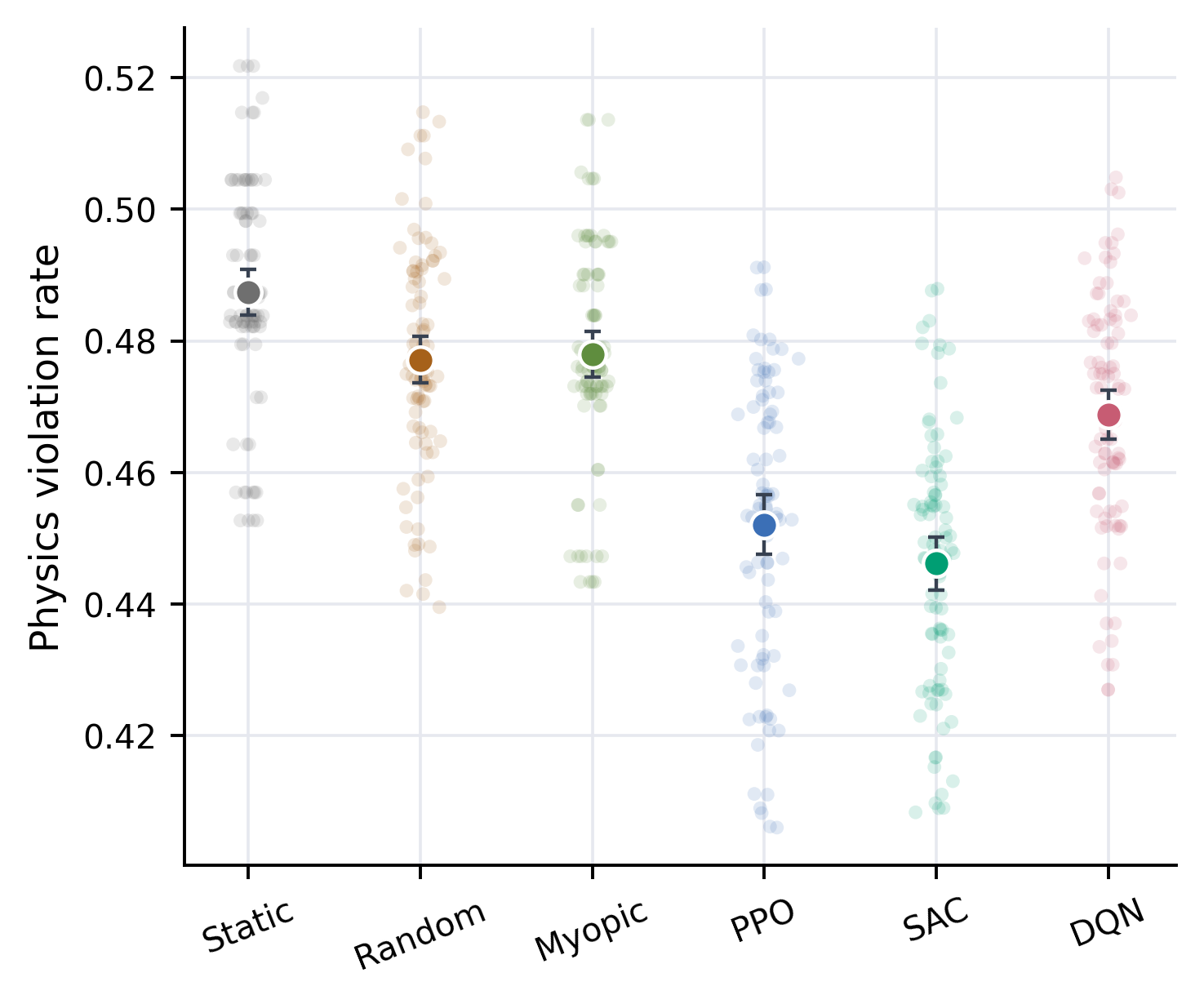}
\caption{Physics violation rate.}
\label{fig:rq1-physics-violation}
\end{subfigure}
\hfill
\begin{subfigure}[t]{0.32\textwidth}
\centering
\includegraphics[width=\linewidth]{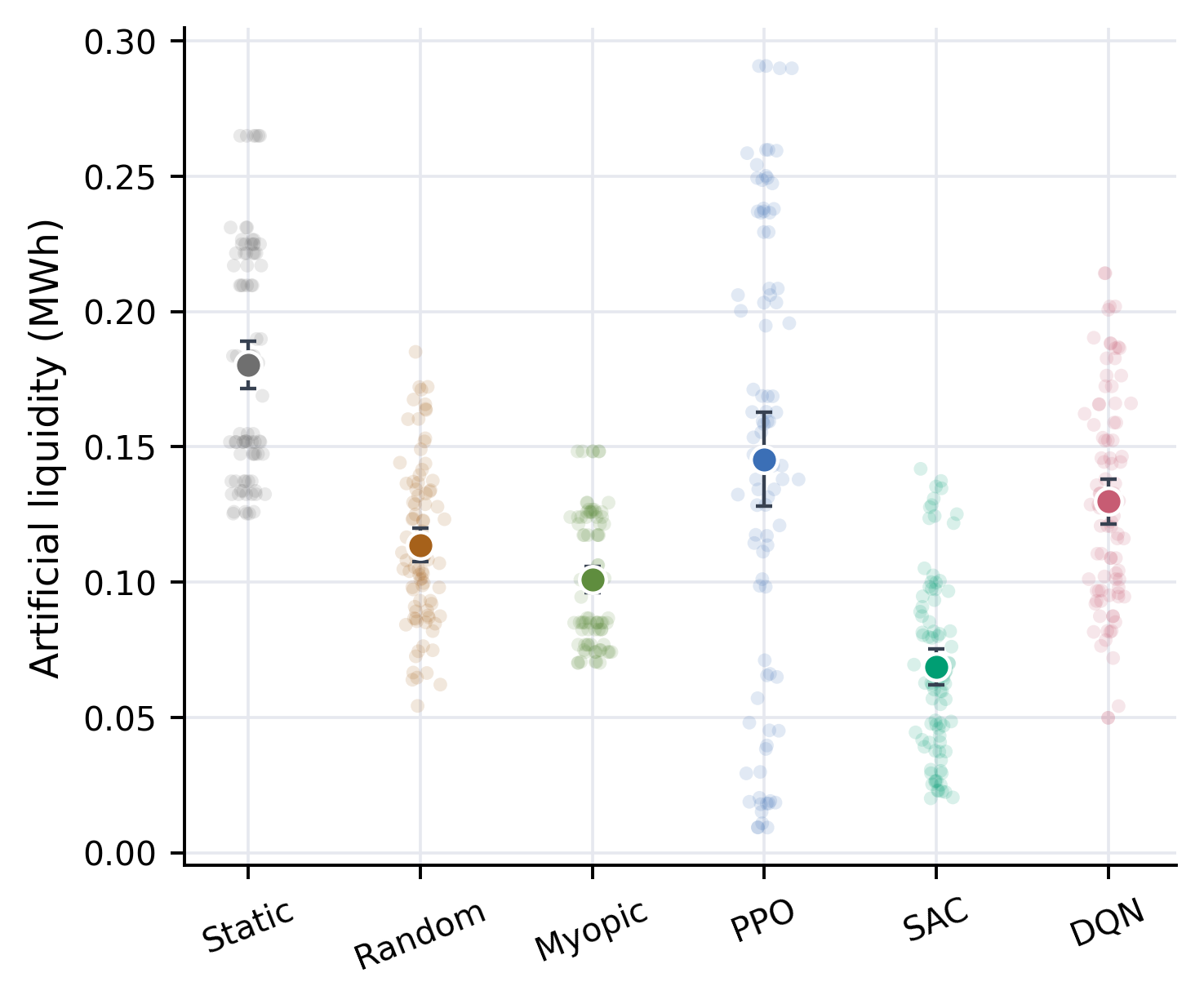}
\caption{Artificial liquidity.}
\label{fig:rq1-artificial-liquidity}
\end{subfigure}
\hfill
\begin{subfigure}[t]{0.32\textwidth}
\centering
\includegraphics[width=\linewidth]{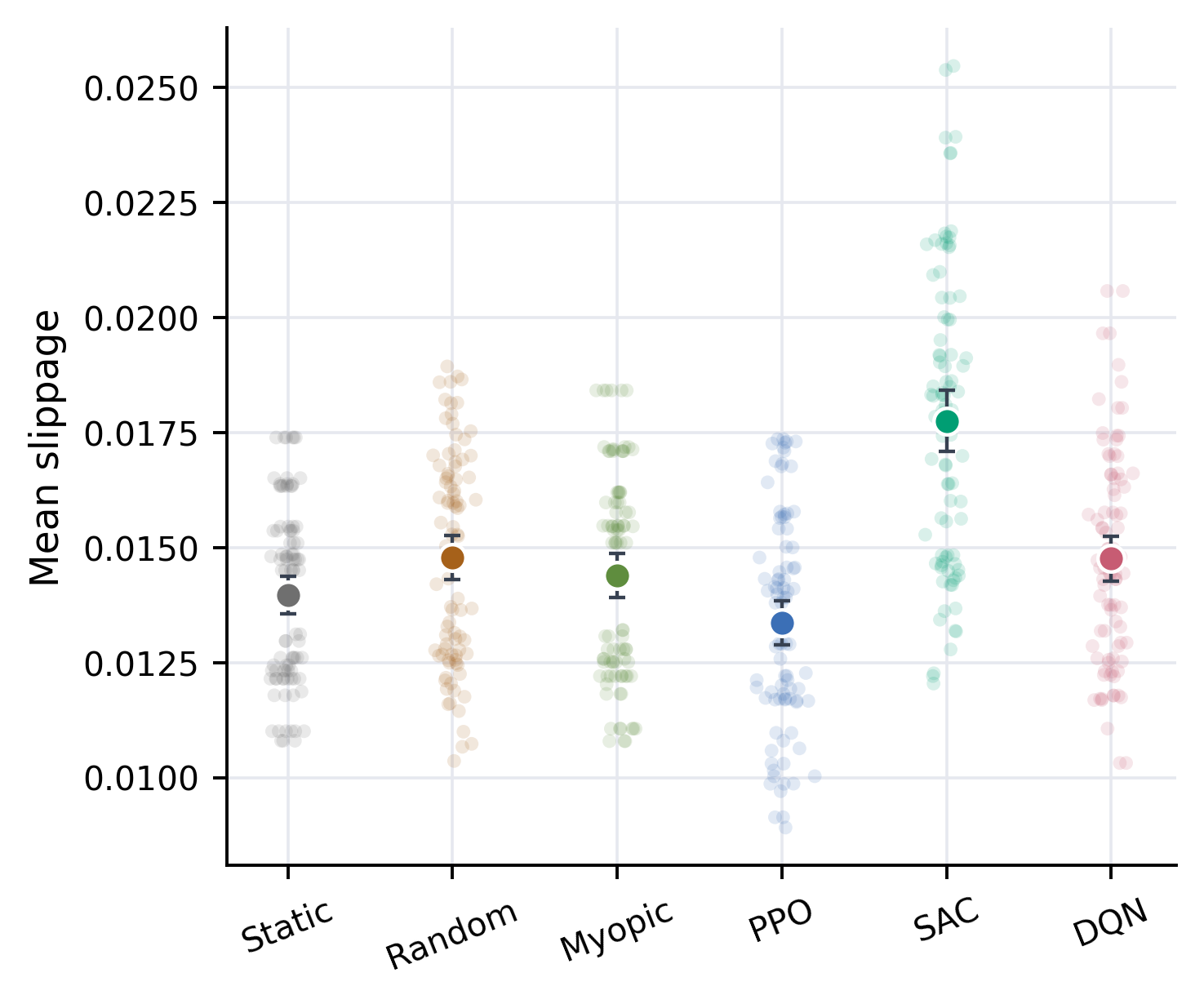}
\caption{Mean slippage.}
\label{fig:rq1-mean-slippage}
\end{subfigure}
\caption{Trustworthiness metrics in the main benchmark. Translucent points are individual rollouts.}
\label{fig:rq1-trustworthiness}
\end{figure*}

\section{Evaluation}
\label{sec:results}

Our empirical evaluation addresses the research questions using data aggregated across three independent seeds. We benchmark three agentic policies (PPO+LLM, SAC+LLM, DQN+LLM) against six non-agentic baselines (PPO, SAC, DQN, Static, Random, Myopic). The  evaluation dataset (Appendix~\ref{app:raw-data-provenance}) comprises 1,620 episode-level records, 38,880 hourly actions, 194,400 hourly state records, and 12,960 LLM governance log entries.

\begin{table*}[t]
\centering
\caption{Paired without-constraints deltas with and without the LLM governance layer. Each row uses 90 paired rollouts across three seeds. Positive $\Delta$ reward indicates that removing the physics penalty increases scalar utility; positive $\Delta$ violation or artificial liquidity indicates increased safety risk.}
\label{tab:physics-ablation-deltas}
\begin{tabular}{llcccc}
\toprule
Governance & Policy & $\Delta$ Reward & $\Delta$ Physics Violation & $\Delta$ Artificial Liquidity & $\Delta$ Slippage \\
\midrule
RL & PPO & 22.61 $\pm$ 1.10 & 0.0311 $\pm$ 0.0084 & 0.0785 $\pm$ 0.0678 & -0.0012 $\pm$ 0.0010 \\
RL & SAC & 22.52 $\pm$ 1.01 & 0.0412 $\pm$ 0.0095 & 0.1238 $\pm$ 0.0386 & -0.0041 $\pm$ 0.0017 \\
RL & DQN & 23.55 $\pm$ 0.91 & 0.0182 $\pm$ 0.0101 & 0.0576 $\pm$ 0.0317 & -0.0018 $\pm$ 0.0014 \\
RL+LLM & PPO & 22.67 $\pm$ 1.17 & 0.0225 $\pm$ 0.0107 & 0.0598 $\pm$ 0.0631 & -0.0011 $\pm$ 0.0016 \\
RL+LLM & SAC & 22.62 $\pm$ 0.96 & 0.0344 $\pm$ 0.0102 & 0.1149 $\pm$ 0.0388 & -0.0038 $\pm$ 0.0017 \\
RL+LLM & DQN & 23.78 $\pm$ 0.88 & 0.0081 $\pm$ 0.0079 & 0.0413 $\pm$ 0.0317 & -0.0015 $\pm$ 0.0018 \\
\bottomrule
\end{tabular}
\end{table*}

\begin{figure*}[t]
\centering
\begin{subfigure}[t]{0.32\textwidth}
\centering
\includegraphics[width=\linewidth]{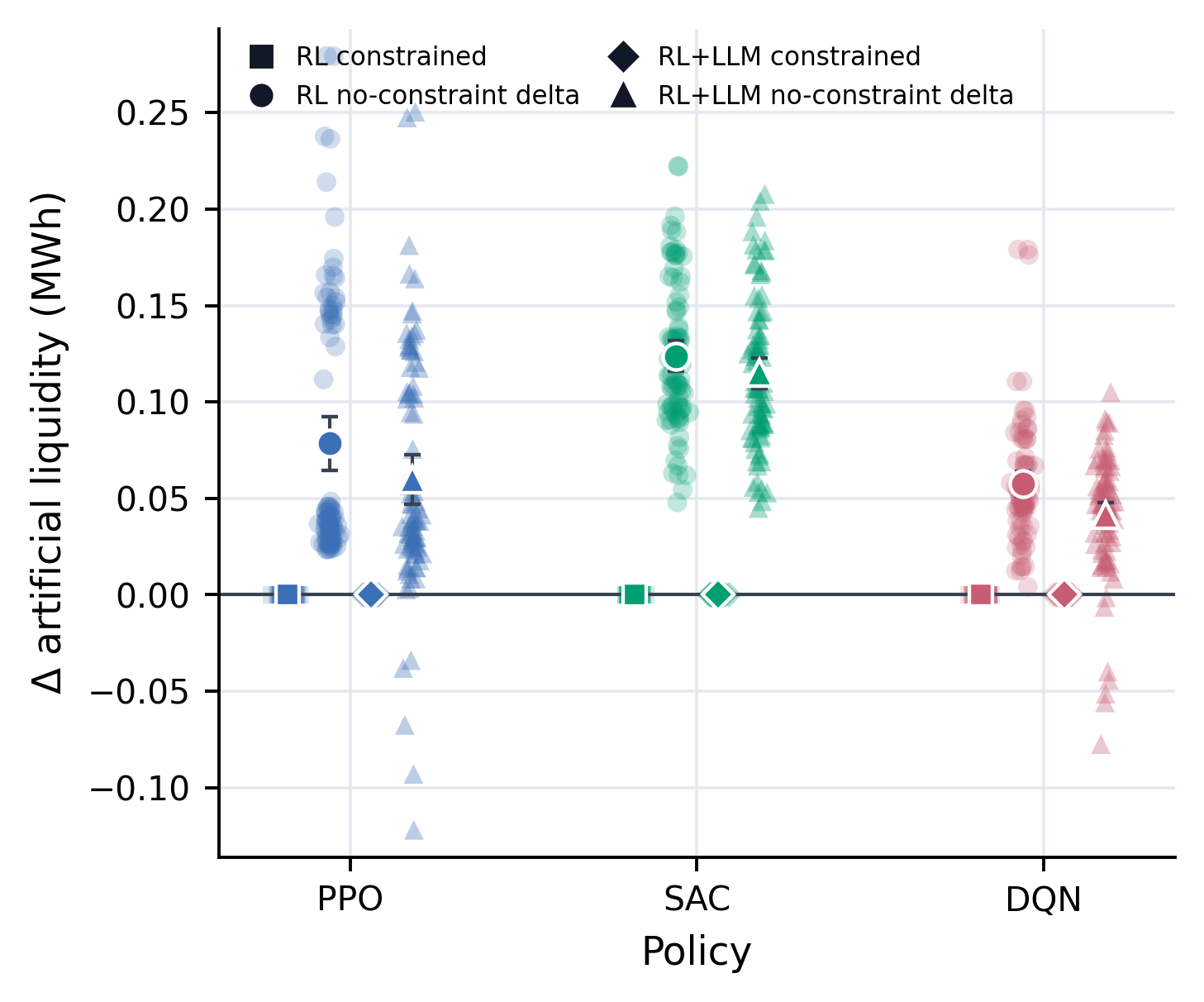}
\caption{Paired without-constraints artificial-liquidity deltas for RL and RL+LLM.}
\label{fig:rq2-paired-delta}
\end{subfigure}
\hfill
\begin{subfigure}[t]{0.32\textwidth}
\centering
\includegraphics[width=\linewidth]{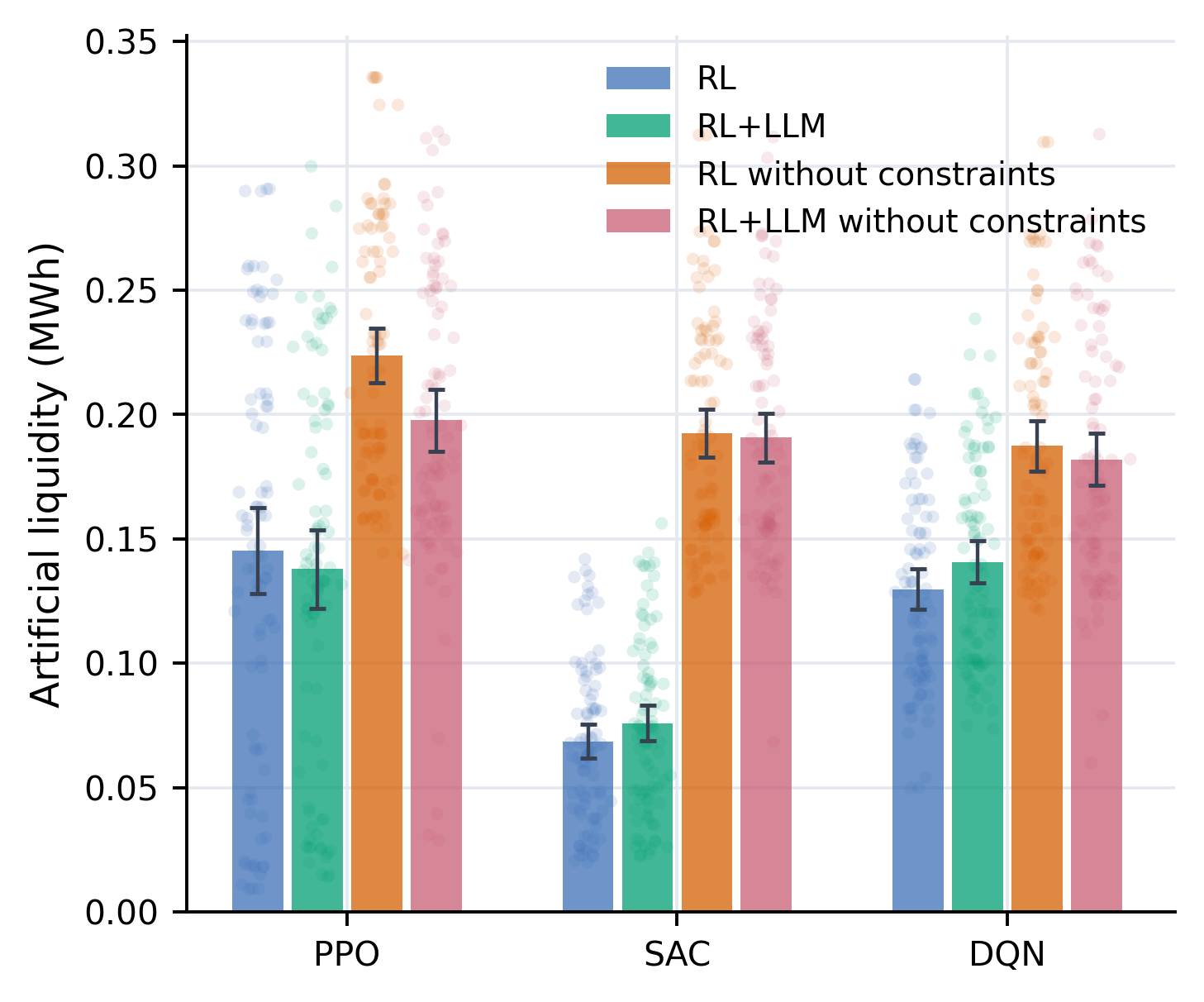}
\caption{Artificial liquidity across RL, RL+LLM, and their without-constraints ablations.}
\label{fig:rq2-liquidity-bars}
\end{subfigure}
\hfill
\begin{subfigure}[t]{0.32\textwidth}
\centering
\includegraphics[width=\linewidth]{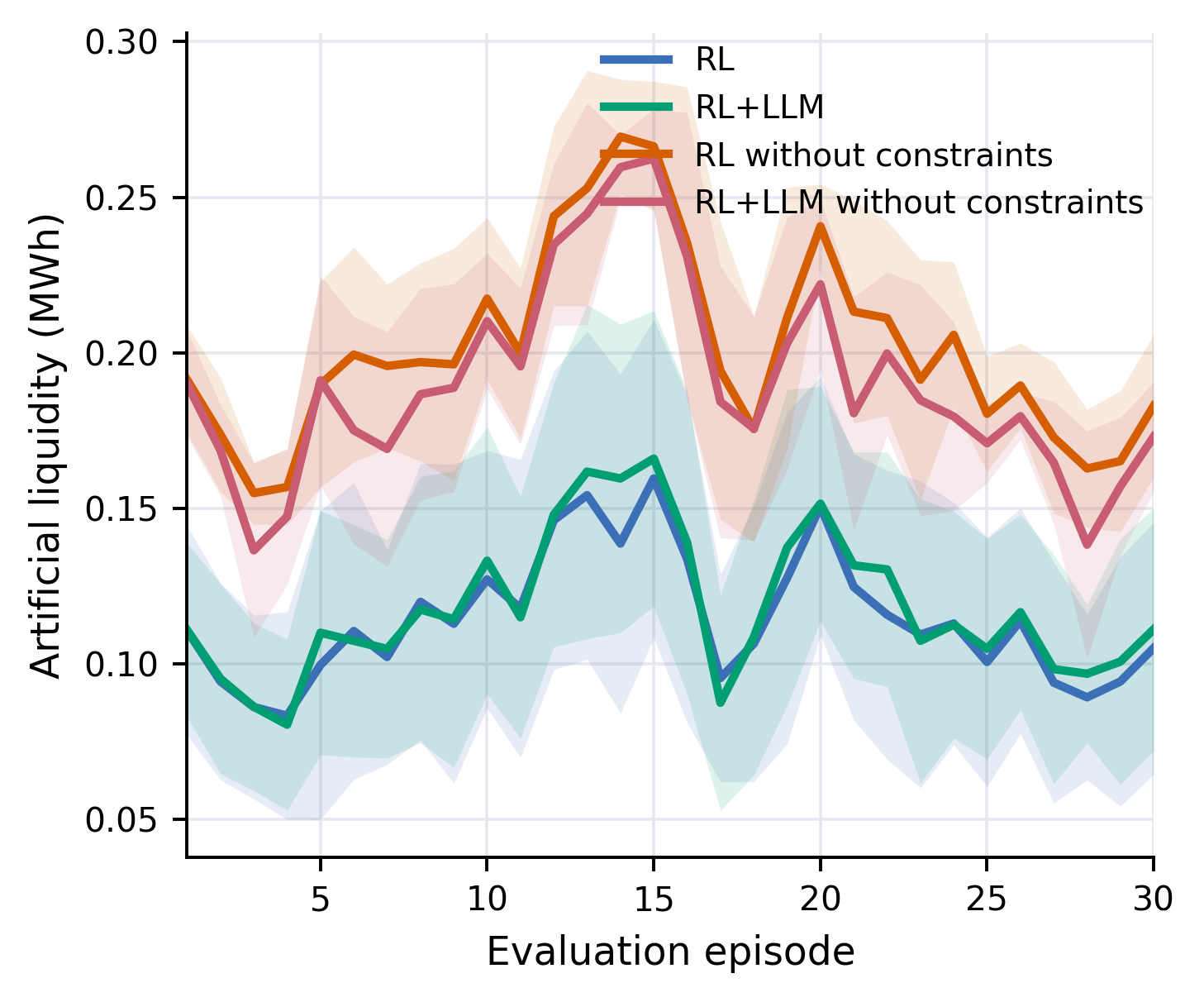}
\caption{Episode-wise artificial-liquidity trajectories for all four learned-agent settings.}
\label{fig:rq2-liquidity-trajectory}
\end{subfigure}
\caption{Ablation experiment of physics constraints. Translucent points are rollout-level or paired-rollout observations.}
\label{fig:rq2-physics-ablation}
\end{figure*}

\subsection{Benchmark Performance}
\label{subsec:rq1}

We first evaluate the RL policies against three baseline models(Static, Random, Myopic) for market governance.  To assess trustworthiness comprehensively, Table~\ref{tab:main-performance} reports utility, physical consistency, and market-execution metrics simultaneously. Results indicate that learned RL control provides a clear advantage over non-RL baselines. Across 90 rollouts per policy, the RL policies achieve a higher mean cumulative reward than the static, random, and myopic baselines ($-22.61$ vs. $-23.97$; Welch's $t$-test, $p<10^{-28}$). This improvement is primarily driven by continuous-control methods, as both PPO and SAC outperform all baselines ($p<10^{-6}$ for all pairwise tests). Notably, these non-RL baselines are non-trivial: they represent stable governance rules (static), pure action-space exploration (random), and local rationality (myopic). Their underperformance suggests that effective market governance requires adaptive RL rather than rigid or one-step heuristics.

However, trustworthiness metrics reveal a complex utility-safety tradeoff. While SAC achieves the highest cumulative reward and trading volume alongside the lowest physics violation rate, it suffers from the highest mean slippage among the RL policies. Conversely, PPO offers the lowest slippage but generates more artificial liquidity. DQN improves upon random control but fails to consistently outperform the myopic baseline, illustrating that RL architecture choice is critical for fine-grained control tasks. Consequently, these patterns expose an evaluation frontier. For rigorous trustworthiness evaluation,  maximizing reward does not guarantee a safe policy if the gain introduces market friction or lacks physical backing. Figure~\ref{fig:rq1-main-benchmark} illustrates this frontier, showing the distinct operational trade-offs and residual variability within each policy type.

The metric decomposition in Figure~\ref{fig:rq1-trustworthiness} further unpacks this frontier. Although RL policies reduce overall physics violations compared to non-RL baselines, the underlying risk is redistributed across artificial liquidity and slippage. SAC excels at physical credibility, whereas PPO prioritizes slippage control. 

\begin{table*}[t]
\centering
\caption{Paired effects of adding the LLM Planner/Auditor layer under the physics-constrained reward. $\Delta$ columns are RL+LLM minus RL over 90 matched rollouts across three seeds; audit and revision rates are computed from 2,160 governance log steps per policy.}
\label{tab:auditability}
\begin{tabular}{lcccccc}
\toprule
Policy & $\Delta$ Reward & $\Delta$ Action Jitter & $\Delta$ Artificial Liquidity & $\Delta$ Spatial Fairness & Audit Rate & Revision Rate \\
\midrule
PPO & -0.094 $\pm$ 0.293 & 0.0223 $\pm$ 0.0154 & -0.0074 $\pm$ 0.0216 & 0.0004 $\pm$ 0.0008 & 0.344 & 0.961 \\
SAC & -0.062 $\pm$ 0.268 & -0.0290 $\pm$ 0.0198 & 0.0073 $\pm$ 0.0110 & 0.0003 $\pm$ 0.0006 & 0.342 & 0.966 \\
DQN & -0.277 $\pm$ 0.503 & -0.0179 $\pm$ 0.0274 & 0.0109 $\pm$ 0.0252 & -0.0002 $\pm$ 0.0021 & 0.372 & 0.958 \\
\bottomrule
\end{tabular}
\end{table*}

\begin{figure*}[t]
\centering
\begin{subfigure}[t]{0.32\textwidth}
\centering
\includegraphics[width=\linewidth]{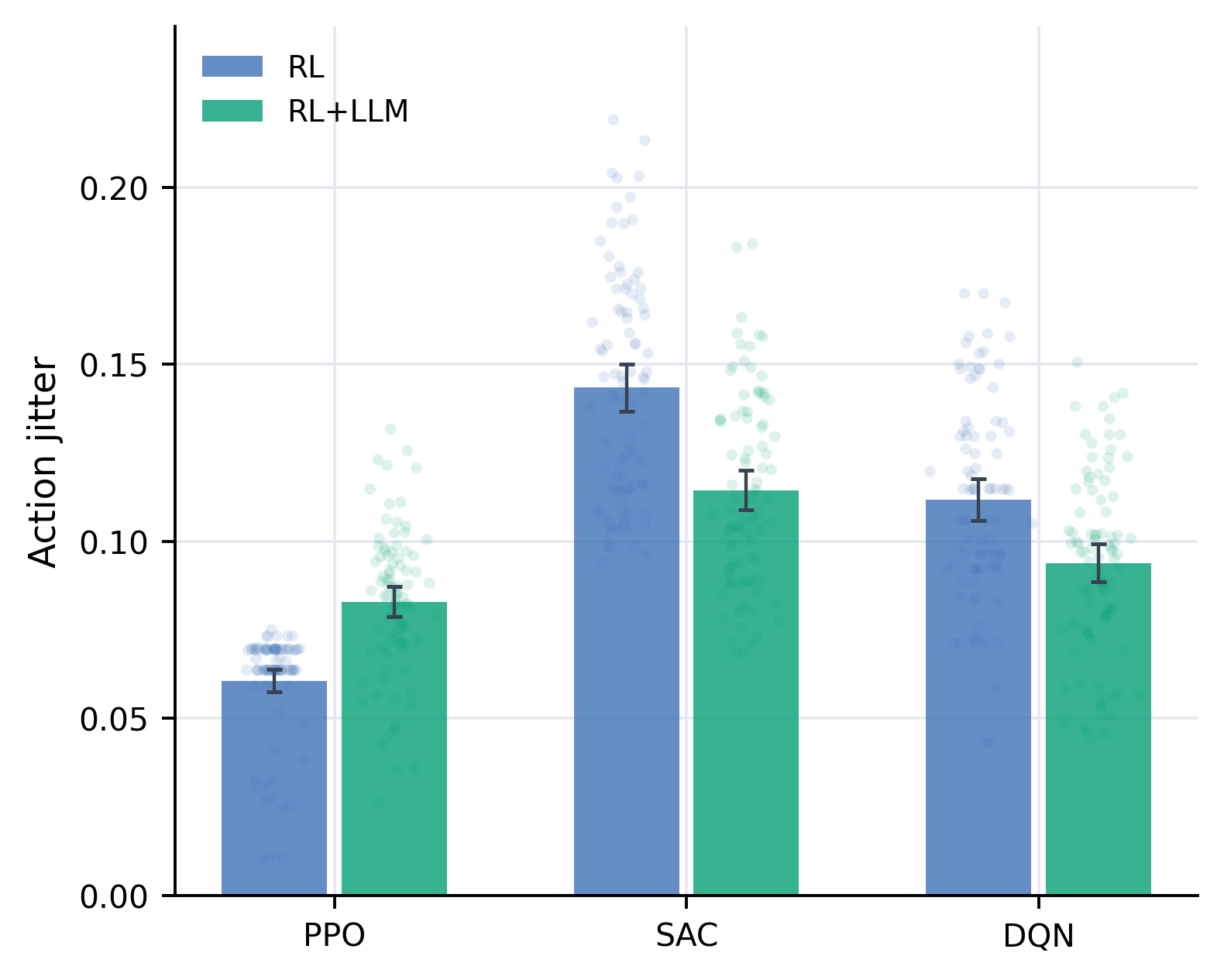}
\caption{Action stability: RL+LLM reduces action jitter for SAC and DQN,}
\label{fig:rq3-audit-rates}
\end{subfigure}
\hfill
\begin{subfigure}[t]{0.32\textwidth}
\centering
\includegraphics[width=\linewidth]{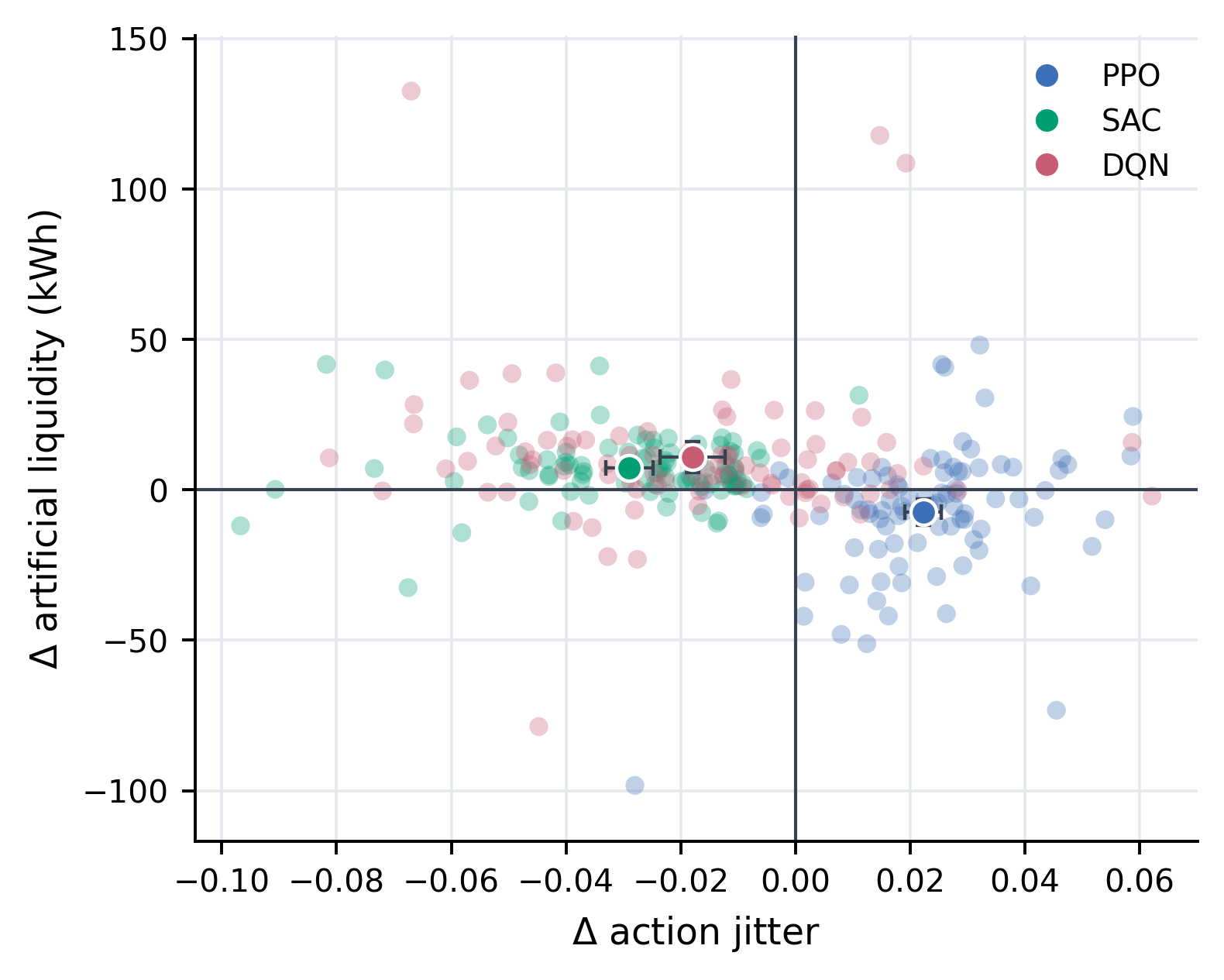}
\caption{Paired RL+LLM-minus-RL changes in action jitter and artificial liquidity.}
\label{fig:rq3-agentic-delta}
\end{subfigure}
\hfill
\begin{subfigure}[t]{0.32\textwidth}
\centering
\includegraphics[width=\linewidth]{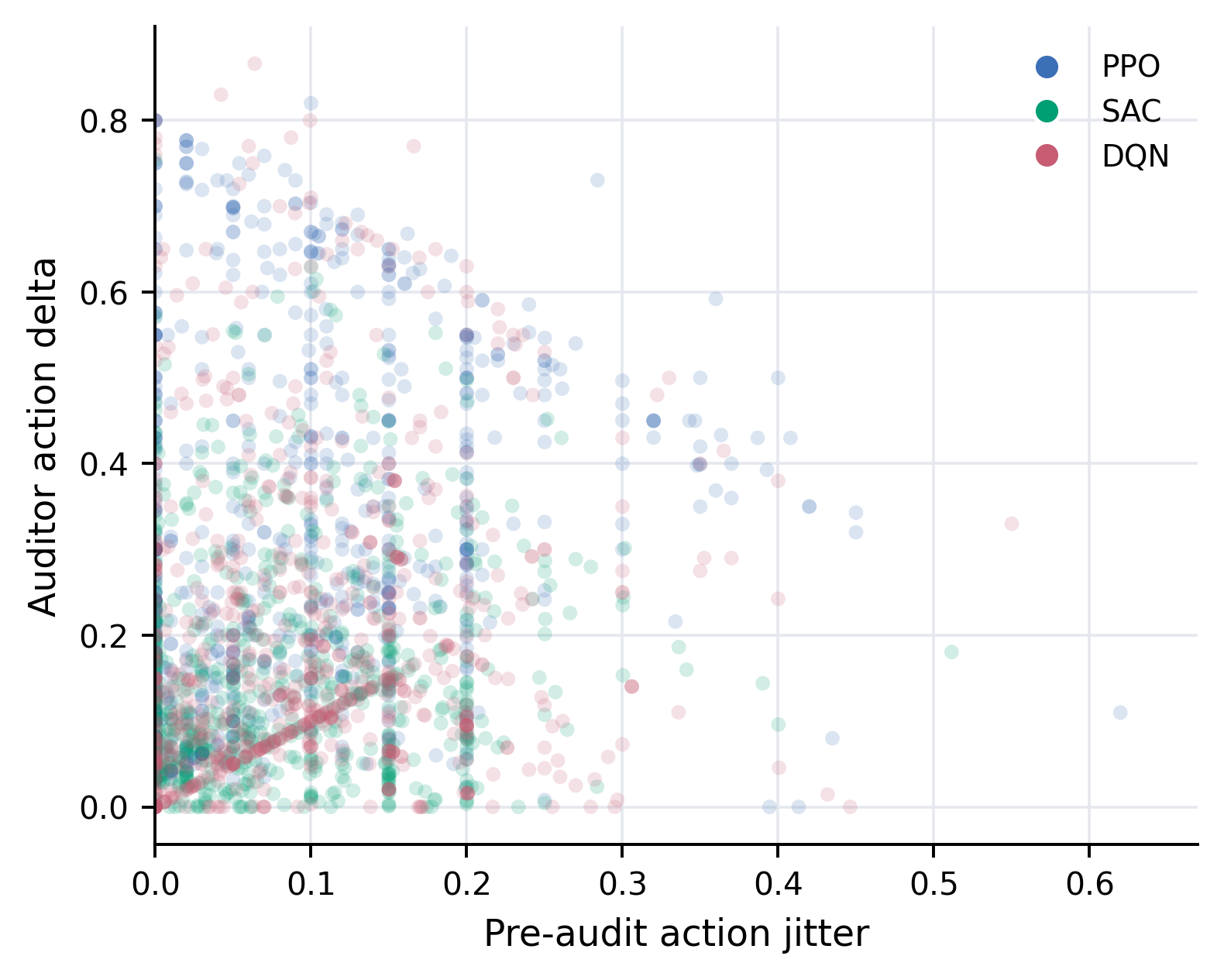}
\caption{Step-level relationship between pre-audit action jitter and Auditor action delta.}
\label{fig:rq3-audit-step-delta}
\end{subfigure}
\caption{Agentic governance diagnostics for the LLM Planner/Auditor layer. The panels combine summary auditability statistics, paired episode-level governance effects, and step-level audit traces.}
\label{fig:rq3-governance}
\end{figure*}

\subsection{Ablation Experiment of Physics Constraints}
\label{subsec:rq2}

The ablation study investigates the necessity of adding physical constraints to our RL and AI agent models. We examine four configurations: standard RL, RL without physics constraints, RL with LLM governance, and RL with LLM governance but lacking physics constraints. In the unconstrained settings, we remove the physics penalty from the reward function(Eq.6) while continuing to track physical violations and artificial liquidity. This design decouples the agent's optimization objective from the benchmark's safety diagnostics. Consequently, scalar reward maximization does not inherently equate to a better market outcome. If an agent boosts its reward by backing invalid supply and inflating apparent liquidity, the benchmark can explicitly expose it.

The empirical results in Table~\ref{tab:physics-ablation-deltas} demonstrate that removing the physics penalty inflates scalar rewards, but at the cost of severe safety risks. Under raw RL deployment, artificial liquidity surges by 0.0785, 0.1238, and 0.0576 MWh for PPO, SAC, and DQN, respectively. This failure mode persists even under RL+LLM deployment, where artificial liquidity still increases by 0.0598, 0.1149, and 0.0413 MWh. This illustrates a consistent shift toward higher artificial liquidity when constraints are ablated across both governance regimes (Figure~\ref{fig:rq2-paired-delta}). Although the LLM layer mitigates the magnitude of these violations, it cannot eliminate the misalignment caused by a misspecified reward. 

This comparison highlights the complementary roles of physical constraints and AI agent governance. While LLM governance audits actions post-hoc, physics-constrained RL embeds safety during optimization to prevent the exploitation of unviable liquidity. Aggregated across learned policies, mean artificial liquidity remains low in constrained environments (0.1146 MWh for RL; 0.1181 MWh for RL+LLM) but spikes in unconstrained settings (0.2012 MWh for RL; 0.1901 MWh for RL+LLM) (Figure~\ref{fig:rq2-liquidity-bars}). Furthermore, this risk inflation is not an artifact of isolated rollouts, but a persistent behavioral trajectory woven throughout the evaluation episodes (Figure~\ref{fig:rq2-liquidity-trajectory}). This confirms that unconstrained optimization remains riskier, even with active agentic oversight.

\subsection{Evaluation of Agentic Layer}
\label{subsec:rq3}

Having proven the function of physics constraints, we now evaluate the AI agent layer's capacity for verifiable oversight. By analyzing transparent audit traces from market execution, we assess the system's effectiveness as an auditable risk-control interface. This evaluation specifically quantifies intervention triggers, the magnitude of action modifications, and the underlying rationales for these corrections.

To investigate the reliability of the AI agent mechanism, Table~\ref{tab:auditability} presents  metrics derived from 12,960 step-level governance logs (encompassing three seeds, two constraint settings, and three learned policies). The system demonstrates exceptional reliability, with a Planner validity of 1.0 across all configurations. Notably, the evaluation reveals that effective oversight is highly selective rather than continuous. Under the physics-constrained setting, the Auditor intervenes in 34.2\%--37.2\% of steps, whereas in the unconstrained setting, the intervention rate is 27.3\%--29.7\%. The Auditor revises proposed actions in over 95\% of cases under constraints, and over 83\% without constraints.

The empirical results suggest that agentic governance functions as an auditable risk-control interface rather than a universal performance optimizer. Its impact is highly dependent on the behavioral characteristics of the base RL policy. For instance(Figure\ref{fig:rq3-audit-rates}), the LLM layer successfully suppresses action jitter for SAC and DQN by 0.0290 and 0.0179, respectively. Conversely, it marginally increases PPO jitter by 0.0223, primarily due to more frequent clipping and targeted revisions of PPO's proposed actions. Similar policy-dependent trade-offs emerge in artificial liquidity: oversight reduces it for PPO ($-0.0074$ MWh) but slightly increases it for SAC ($+0.0073$ MWh) and DQN ($+0.0109$ MWh). Spatial fairness remains largely stable, with minimal deltas across all policies(Figure~\ref{fig:rq3-agentic-delta}).

\begin{table}[t]
\centering
\small
\caption{Representative LLM audit trace selected from \texttt{agentic\_logs.jsonl}. The row records the market context, proposed RL action, audited final action, and action delta.}
\label{tab:app-case-study}
\begin{tabularx}{\columnwidth}{@{}p{0.31\columnwidth}X@{}}
\toprule
Field & Value \\
\midrule
policy / setting &  PPO / LLM No-physics \\
Episode and step & Episode 6, step 15, hour 15 \\
Trigger context & raw violation 0.14; gap 1.97; static slippage 0.72 \\
RL proposed action & r=0.049, l=0.931, b=0.200 \\
Audited final action & r=0.050, l=0.200, b=0.000 \\
Action delta & 0.932 \\
Auditor rationale & Violation rate exceeds audit threshold; tighten allocation to limit exposure to unsafe backed supply. \\
\bottomrule
\end{tabularx}
\end{table}

These mixed outcomes reveal that researchers should not assume that deploying an LLM governance layer automatically resolves all safety risks. Instead, it must rigorously define the boundaries of such oversight. Within the decentralized energy market setting, \proj demonstrates that while the AI agent successfully intercepts unstable decisions, as evidenced by the positive correlation between pre-audit action jitter and the corrective delta in Figure~\ref{fig:rq3-audit-step-delta}, it cannot fully compensate for a  misspecified RL reward. Ultimately, our findings advocate for a complementary approach to trustworthy autonomous systems. While agentic governance guarantees post-deployment transparency and event-driven risk mitigation, physically constrained RL optimization remains necessary to prevent reward-driven exploitation.

\subsection{Case Study: Transparent LLM Audit Trace}
\label{subsec:case-study}

\begin{figure}[t]
    \centering
    \begin{subfigure}[t]{0.48\linewidth}
        \centering
        \includegraphics[width=\linewidth]{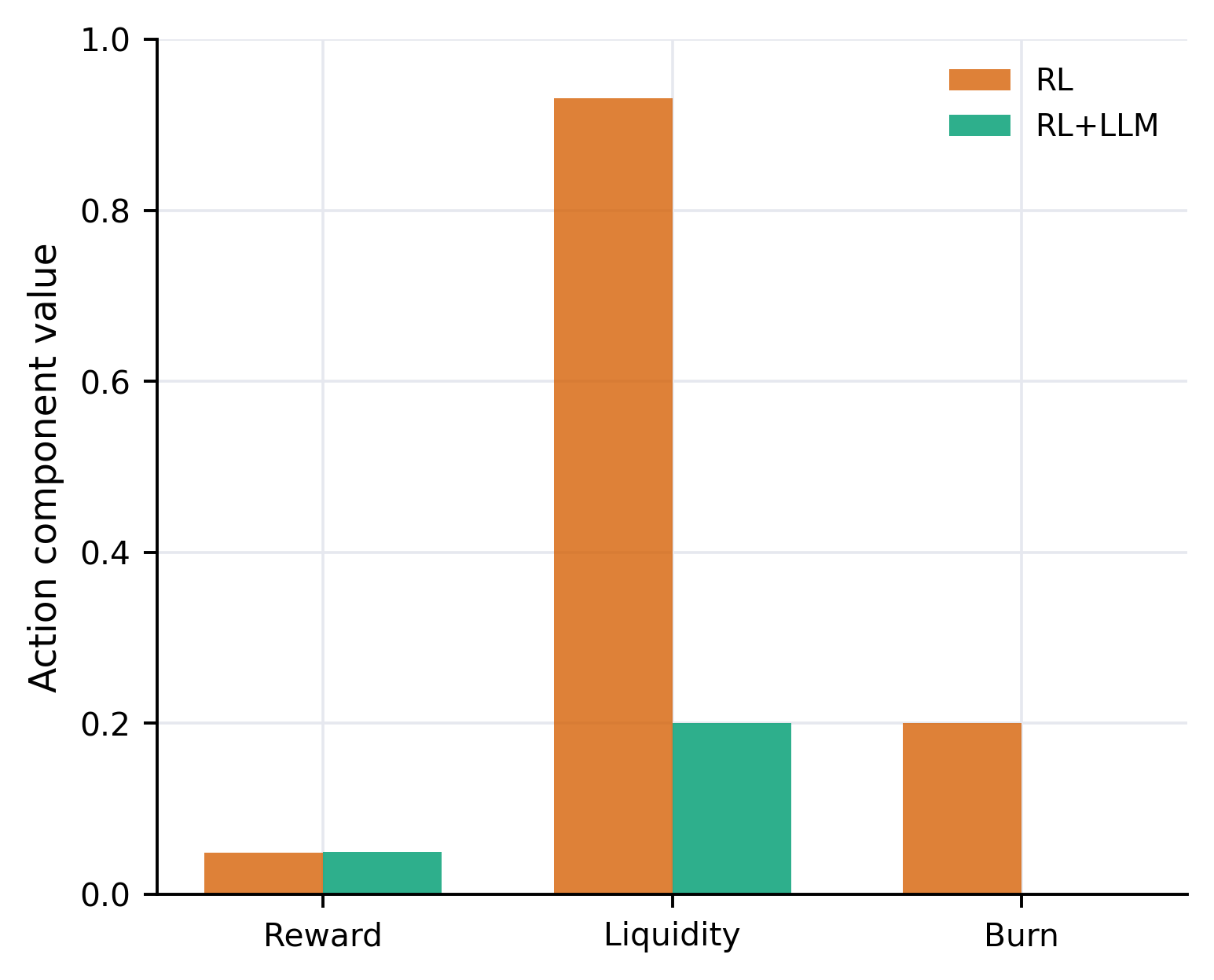}
        \caption{Selected action revision.}
        \label{fig:rq4_action_revision}
    \end{subfigure}
    \hfill
     \begin{subfigure}[t]{0.48\linewidth}
        \centering
        \includegraphics[width=\linewidth]{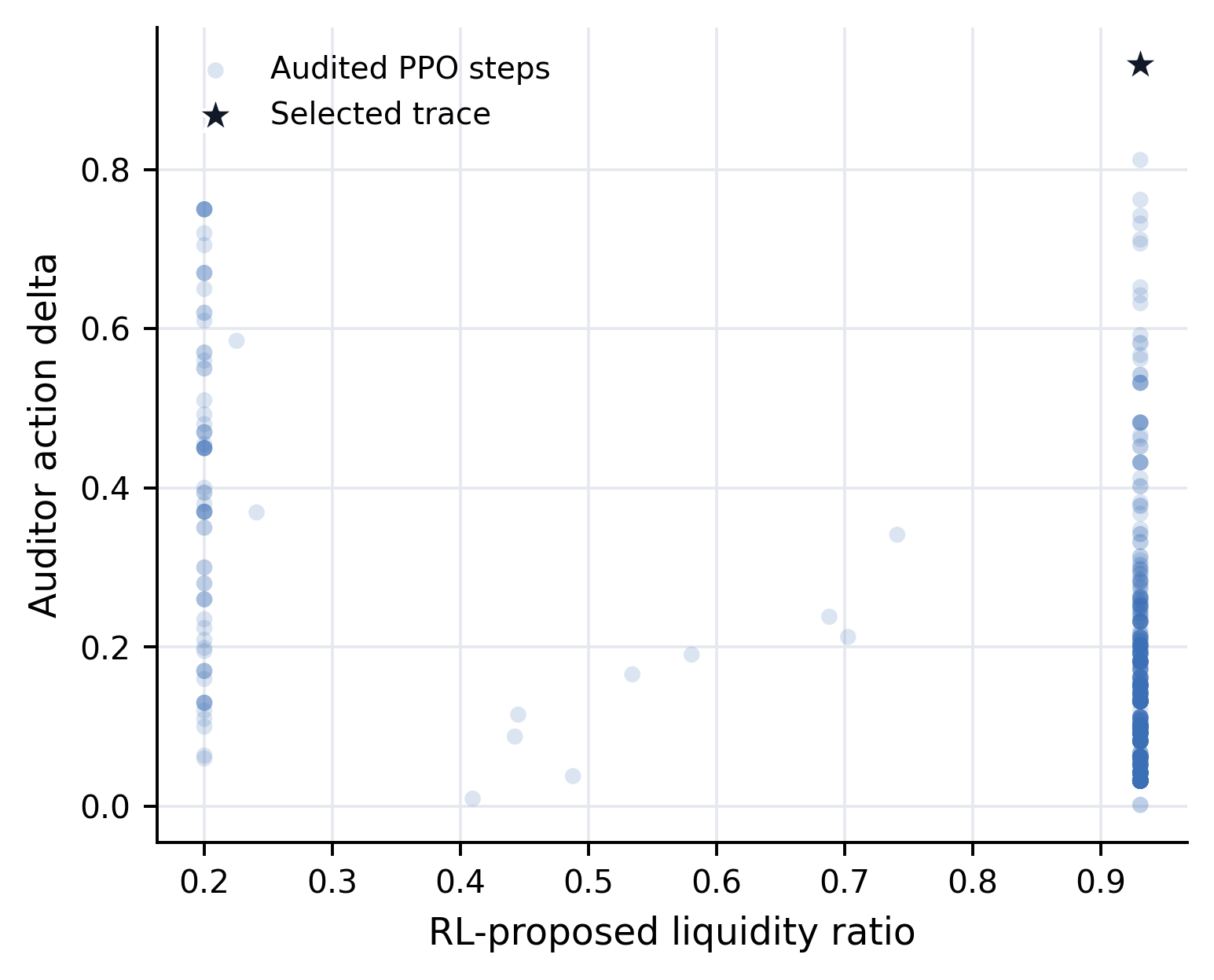}
        \caption{Context of audited PPO steps.}
        \label{fig:rq4_trace_context}
    \end{subfigure}
    \caption{Case-study audit trace for PPO under the LLM governance setting.}
    \label{fig:rq4_audit_trace}
\end{figure}

To illustrate the transparency of our evaluation, we analyze a representative audit trace (Table~\ref{tab:app-case-study}). We select a scenario where the PPO agent operates without physical constraints, a setting where reward misspecification inherently encourages risky market behavior. At the evaluated step, environmental risk signals are  elevated, including a raw physics violation rate of 0.14 and a supply gap of 1.97.

Driven by economic utility, the RL policy proposes an aggressive, high-liquidity action $(r=0.049, l=0.931, b=0.200)$. Recognizing the safety breach, the LLM Auditor intervenes and revises the action to a conservative bound $(r=0.050, l=0.200, b=0.000)$, resulting in a substantial action delta of 0.932 (Figure \ref{fig:rq4_trace_context}). Beyond the numerical correction, the physical violation rate exceeded the safety threshold, necessitating a tighter liquidity allocation to limit exposure to unsafe supply.

This case validates the  contribution of the agentic layer to trustworthy evaluation. \proj records the entire decision-making process. It explicitly documents the activating risk signals, the resulting action adjustments, and the auditor's natural language rationale. By providing this  transparency,  \proj guarantees that autonomous interventions remain accountable after deployment.

\section{Conclusion}
In conclusion, we proposed \proj, a physics-constrained benchmark designed to evaluate the trustworthiness of economic agents in decentralized energy markets. Our empirical evaluations reveal a clear trade-off inherent in autonomous market governance. While RL policies improve economic utility relative to static and heuristic rules, they may compromise safety for market efficiency. This vulnerability is starkly exposed when physical constraints are removed: driven purely by reward maximization, an RL policy may exploit false data injection attacks to artificially inflate market liquidity. To counter this flaw, combining physical constraints with an AI-agent governance layer provides a structured and auditable safety interface. By evaluating context and intercepting high-risk decisions, the AI-agent layer creates transparent intervention traces and can mitigate instability patterns, although its corrective impact remains policy-dependent and bounded by the underlying RL controller.

The deployment of autonomous agents in cyber-physical systems requires evaluation paradigms that look beyond standard reward maximization. \proj addresses this gap by logging the complete decision loop, including environmental risk signals, action corrections, and natural language explanations. Consequently, our benchmark supports deployment-oriented stress tests in which agent behaviors remain economically viable, verifiable, accountable, and physically grounded.

Future work will make \proj broader and more diagnostic. We will test whether the utility--safety--stability--fairness--auditability template transfers to autonomous trading and supply-chain allocation by replacing the PV physics oracle with domain-specific feasibility or risk checks. We will also calibrate the simulator with broader meteorological and market traces and run sensitivity analyses over reward weights, FDIA frequency and severity, node scale, and market-clearing assumptions. For the Planner/Auditor, we will measure selectivity with safe-action holdouts, approval-only controls, false-positive and false-negative rates, and trigger-level ablations, separating meaningful risk detection from overly conservative defaults. Finally, we will compare \proj with constrained-RL and agentic-benchmark baselines to clarify when post-hoc governance helps and when reward misspecification must be addressed during training.

\bibliographystyle{ACM-Reference-Format}
\newpage
\bibliography{references}

@article{gu2024review,
  author={Gu, Shangding and others},
  title={A Review of Safe Reinforcement Learning: Methods, Theory and Applications},
  journal={IEEE Transactions on Pattern Analysis and Machine Intelligence},
  year={2024},
  publisher={IEEE}
}

@article{chen2022peer,
  author={Chen, T. and Bu, S. and Liu, X. and Kang, J. and Yu, F. R. and Han, Z.},
  title={Peer-to-peer energy trading and energy conversion in interconnected multi-energy microgrids using multi-agent deep reinforcement learning},
  journal={IEEE Transactions on Smart Grid},
  volume={13},
  number={1},
  pages={715--727},
  year={2022},
  publisher={IEEE}
}

@article{malinova2023tokenomics,
  title={Tokenomics: When Tokens Beat Equity},
  author={Malinova, Katya and Park, Andreas},
  journal={Management Science},
  volume={69},
  number={11},
  pages={6568--6583},
  year={2023},
  publisher={INFORMS}
}

@article{cong2021tokenomics,
  title={Tokenomics: Dynamic adoption and valuation},
  author={Cong, Lin William and Li, Ye and Wang, Neng},
  journal={The Review of Financial Studies},
  volume={34},
  number={3},
  pages={1105--1155},
  year={2021},
  publisher={Oxford University Press}
}

@article{raffin2021stable,
  author  = {Raffin, Antonin and Hill, Ashley and Gleave, Adam and Kanervisto, Anssi and Ernestus, Maximilian and Dormann, Noah},
  title   = {Stable-Baselines3: Reliable Reinforcement Learning Implementations},
  journal = {Journal of Machine Learning Research},
  year    = {2021},
  volume  = {22},
  number  = {268},
  pages   = {1--8}
}

@article{schulman2017proximal,
  title={Proximal policy optimization algorithms},
  author={Schulman, John and Wolski, Filip and Dhariwal, Prafulla and Radford, Alec and Klimov, Oleg},
  journal={arXiv preprint arXiv:1707.06347},
  year={2017}
}

@inproceedings{haarnoja2018soft,
  title={Soft actor-critic: Off-policy maximum entropy deep reinforcement learning with a stochastic actor},
  author={Haarnoja, Tuomas and Zhou, Aurick and Abbeel, Pieter and Levine, Sergey},
  booktitle={International conference on machine learning},
  pages={1861--1870},
  year={2018},
  organization={PMLR}
}

@article{amiri2025deep,
  title={Deep Reinforcement Learning with Local Interpretability for Transparent Microgrid Resilience Energy Management},
  author={Amiri, Mohammad Hossein Nejati and Annaz, Fawaz and De Oliveira, Mario and Gueniat, Florimond},
  journal={arXiv preprint arXiv:2508.08132},
  year={2025}
}

@inproceedings{lozano2025democratizing,
  title={Democratizing microgrid optimization: An llm agent for dispatching mobile chargers to construction electric vehicles},
  author={Lozano, Daniela Rojas and Shi, Yuanyuan},
  booktitle={NeurIPS 2025 Workshop on Bridging Language, Agent, and World Models for Reasoning and Planning},
  year={2025}
}

@article{yao2025reward,
  title={Reward Evolution with Graph-of-Thoughts: A Bi-Level Language Model Framework for Reinforcement Learning},
  author={Yao, Changwei and Liu, Xinzi and Li, Chen and Savvides, Marios},
  journal={arXiv preprint arXiv:2509.16136},
  year={2025}
}

@article{yang2025rl2,
  title={RL2: Reinforce Large Language Model to Assist Safe Reinforcement Learning for Energy Management of Active Distribution Networks},
  author={Yang, Xu and Lin, Chenhui and Liu, Haotian and Wu, Wenchuan},
  journal={IEEE Transactions on Smart Grid},
  year={2025}
}

@inproceedings{van2016deep,
  title={Deep reinforcement learning with double q-learning},
  author={Van Hasselt, Hado and Guez, Arthur and Silver, David},
  booktitle={Proceedings of the AAAI conference on artificial intelligence},
  volume={30},
  number={1},
  year={2016}
}

@article{mnih2015human,
  title={Human-level control through deep reinforcement learning},
  author={Mnih, Volodymyr and Kavukcuoglu, Koray and Silver, David and Rusu, Andrei A and Veness, Joel and Bellemare, Marc G and Graves, Alex and Riedmiller, Martin and Fidjeland, Andreas K and Ostrovski, Georg and others},
  journal={nature},
  volume={518},
  number={7540},
  pages={529--533},
  year={2015},
  publisher={Nature Publishing Group}
}

@article{yang2025llm,
  title={LLM-powered distributed optimal scheduling for industrial heat-electricity micro-grids},
  author={Yang, Haolan and Li, Zhengbo and Liu, Youbo and Xiang, Yue and Li, Lingtao and Yang, Jianping and Tan, Ling and Wang, Shiqian and Ma, Huangqi and Xi, Zirui and others},
  journal={IEEE Transactions on Industry Applications},
  year={2025},
  publisher={IEEE}
}

@inproceedings{du2023guiding,
  title={Guiding pretraining in reinforcement learning with large language models},
  author={Du, Yuqing and Watkins, Olivia and Wang, Zihan and Colas, C{\'e}dric and Darrell, Trevor and Abbeel, Pieter and Gupta, Abhishek and Andreas, Jacob},
  booktitle={International Conference on Machine Learning},
  pages={8657--8677},
  year={2023},
  organization={PMLR}
}

@article{yao2022react,
  title={React: Synergizing reasoning and acting in language models},
  author={Yao, Shunyu and Zhao, Jeffrey and Yu, Dian and Du, Nan and Shafran, Izhak and Narasimhan, Karthik and Cao, Yuan},
  journal={arXiv preprint arXiv:2210.03629},
  year={2022}
}

@article{rudin2019stop,
  title={Stop explaining black box machine learning models for high stakes decisions and use interpretable models instead},
  author={Rudin, Cynthia},
  journal={Nature machine intelligence},
  volume={1},
  number={5},
  pages={206--215},
  year={2019},
  publisher={Nature Publishing Group UK London}
}

@article{campos2022soft,
  title={Soft actor-critic deep reinforcement learning with hybrid mixed-integer actions for demand responsive scheduling of energy systems},
  author={Campos, Gustavo and El-Farra, Nael H and Palazoglu, Ahmet},
  journal={Industrial \& Engineering Chemistry Research},
  volume={61},
  number={24},
  pages={8443--8461},
  year={2022},
  publisher={ACS Publications}
}

@inproceedings{carta2023grounding,
  title={Grounding large language models in interactive environments with online reinforcement learning},
  author={Carta, Thomas and Romac, Cl{\'e}ment and Wolf, Thomas and Lamprier, Sylvain and Sigaud, Olivier and Oudeyer, Pierre-Yves},
  booktitle={International Conference on Machine Learning},
  pages={3676--3713},
  year={2023},
  organization={PMLR}
}

@article{dulac2021challenges,
  title={Challenges of real-world reinforcement learning: definitions, benchmarks and analysis},
  author={Dulac-Arnold, Gabriel and Levine, Nir and Mankowitz, Daniel J and Li, Jerry and Paduraru, Cosmin and Hester, Todd and others},
  journal={Machine Learning},
  volume={110},
  number={9},
  pages={2419--2468},
  year={2021},
  publisher={Springer}
}

@inproceedings{kelly2020rl,
  title={Reinforcement learning for electricity network operation},
  author={Kelly, Adrian and O'Sullivan, Aidan and de Mars, Patrick and Marot, Antoine},
  booktitle={Electric Power Systems Research},
  volume={189},
  pages={106740},
  year={2020},
  publisher={Elsevier}
}

@article{liu2011false,
  title={False data injection attacks against state estimation in electric power grids},
  author={Liu, Yao and Ning, Peng and Reiter, Michael K},
  journal={ACM Transactions on Information and System Security (TISSEC)},
  volume={14},
  number={1},
  pages={1--33},
  year={2011},
  publisher={ACM New York, NY, USA}
}

@inproceedings{kwon2023reward,
  title={Reward design with language models},
  author={Kwon, Minae and Xie, Sang Michael and Bullard, Kalesha and Sadigh, Dorsa},
  booktitle={The Eleventh International Conference on Learning Representations},
  year={2023}
}

@inproceedings{ma2023eureka,
  title={Eureka: Human-level reward design via coding large language models},
  author={Ma, Yecheng Jason and Liang, William and Wang, Guanzhi and Huang, De-An and Bastani, Osbert and Jayaraman, Dinesh and Zhu, Yuke and Fan, Linxi and Anandkumar, Anima},
  booktitle={The Twelfth International Conference on Learning Representations},
  year={2023}
}

@inproceedings{qu2025rules,
  title={From Rules to Rewards: Reinforcement Learning for Interest Rate Adjustment in DeFi Lending},
  author={Qu, Hanxiao and Gogol, Krzysztof M and Gr{\"o}tschla, Florian and Tessone, Claudio J},
  booktitle={The International Conference on Mathematical Research for Blockchain Economy},
  pages={85--120},
  year={2025},
  organization={Springer}
}

@article{xu2025auto,
  title={Auto. gov: learning-based governance for decentralized finance (DeFi)},
  author={Xu, Jiahua and Feng, Yebo and Perez, Daniel and Livshits, Benjamin},
  journal={IEEE Transactions on Services Computing},
  year={2025},
  publisher={IEEE}
}

@article{zheng2022ai,
  title={The AI Economist: Taxation policy design via two-level deep multiagent reinforcement learning},
  author={Zheng, Stephan and Trott, Alexander and Srinivasa, Sunil and Parkes, David C and Socher, Richard},
  journal={Science advances},
  volume={8},
  number={18},
  pages={eabk2607},
  year={2022},
  publisher={American Association for the Advancement of Science}
}

@article{zhang2018review,
  title={Review on the research and practice of deep learning and reinforcement learning in smart grids},
  author={Zhang, Dongxia and Han, Xiaoqing and Deng, Chunyu},
  journal={CSEE Journal of Power and Energy Systems},
  volume={4},
  number={3},
  pages={362--370},
  year={2018},
  publisher={CSEE}
}

@article{mengelkamp2018designing,
  title={Designing microgrid energy markets: A case study: The Brooklyn Microgrid},
  author={Mengelkamp, Esther and G{\"a}rttner, Johannes and Rock, Kerstin and Kessler, Scott and Orsini, Lawrence and Weinhardt, Christof},
  journal={Applied Energy},
  volume={210},
  pages={870--880},
  year={2018},
  publisher={Elsevier}
}

@article{tushar2018transforming,
  title={Transforming energy networks via peer-to-peer energy trading: The potential of game-theoretic approaches},
  author={Tushar, Wayes and Yuen, Chau and Mohsenian-Rad, Hamed and Saha, Tapan and Poor, H Vincent and Wood, Kristin L},
  journal={IEEE Signal Processing Magazine},
  volume={35},
  number={4},
  pages={90--111},
  year={2018},
  publisher={IEEE}
}

@inproceedings{achiam2017constrained,
  title={Constrained policy optimization},
  author={Achiam, Joshua and Held, David and Tamar, Aviv and Abbeel, Pieter},
  booktitle={International conference on machine learning},
  pages={22--31},
  year={2017},
  organization={PMLR}
}

@article{xi2025rise,
  title={The rise and potential of large language model based agents: A survey},
  author={Xi, Zhiheng and Chen, Wenxiang and Guo, Xin and He, Wei and Ding, Yiwen and Hong, Boyang and Zhang, Ming and Wang, Junzhe and Jin, Senjie and Zhou, Enyu and others},
  journal={Science China Information Sciences},
  volume={68},
  number={2},
  pages={121101},
  year={2025},
  publisher={Springer}
}

@article{vyas2025autonomous,
  title={Autonomous industrial control using an agentic framework with large language models},
  author={Vyas, Javal and Mercang{\"o}z, Mehmet},
  journal={IFAC-PapersOnLine},
  volume={59},
  number={6},
  pages={349--354},
  year={2025},
  publisher={Elsevier}
}

@techreport{nist2023airmf,
  title={Artificial Intelligence Risk Management Framework (AI RMF 1.0)},
  author={{National Institute of Standards and Technology}},
  institution={National Institute of Standards and Technology},
  year={2023},
  number={NIST AI 100-1},
  doi={10.6028/NIST.AI.100-1}
}

@inproceedings{drgona2025safe,
  title={Safe Physics-informed Machine Learning for Dynamics and Control},
  author={Drgona, Jan and Nghiem, Truong X. and Beckers, Thomas and Fazlyab, Mahyar and Mallada, Enrique and Jones, Colin and Vrabie, Draguna and Brunton, Steven L. and Findeisen, Rolf},
  booktitle={Proceedings of the American Control Conference},
  year={2025}
}

@inproceedings{phiri2025auditable,
  title={Creating Characteristically Auditable Agentic AI Systems},
  author={Phiri, Charles Chimwemwe},
  booktitle={Proceedings of the Intelligent Robotics FAIR 2025},
  year={2025},
  publisher={ACM},
  doi={10.1145/3759355.3759356}
}

@article{ou2026solarchain,
  title={SolarChain: Bridging Physical Law, Verifiable Trust, and Sustainable Markets for Urban Energy Resilience},
  author={Ou, Shilin and Xu, Yifan and Zhang, Zhenshan and Zhang, Luyao and Huang, Ming-Chun},
  journal={arXiv preprint arXiv:2605.23162},
  year={2026}
}

\newpage
\appendix
\section{Formulas Implementation}
\label{app:formulas-implementation}

This appendix lists some important equations used by the released \proj code but not separately expanded in the methodology section. 

\paragraph{PV capacity and trust labels.}
The data generator converts weather into a physical PV upper bound and the loader derives the trust labels used by the benchmark:
\begin{equation}
\begin{aligned}
I_{c,t}
&=\mathbb{I}[z_{c,t}<90^\circ]\,
\min\{\max(S_{c,t},0),\,1.08\max(C_{c,t},0)\},\\
\eta^{temp}_{i,t}
&=\mathrm{clip}\{1+\gamma_i(T_{c,t}-25),\,0.78,\,1.08\},\\
P^{max}_{i,t}
&=\max\{0,I_{c,t}A_i\eta_i\eta^{temp}_{i,t}\},\\
G^{v}_{i,t}
&=\mathbb{I}[\mathrm{status}_{i,t}=\mathrm{verified}]P^{reported}_{i,t},\\
X^{excess}_{i,t}
&=\max(P^{reported}_{i,t}-P^{max}_{i,t},0),\\
\mathrm{viol}_{i,t}
&=\mathbb{I}[X^{excess}_{i,t}>10^{-6}\vee \mathrm{FDIA}_{i,t}],\\
R^{rejected}_{i,t}
&=\mathrm{viol}_{i,t}P^{reported}_{i,t}.
\end{aligned}
\end{equation}

\paragraph{Action decoding and allocation budget.}
Continuous policy outputs are decoded into governance ratios and then sanitized against the global allocation budget:
\begin{equation}
\begin{aligned}
\alpha_t
&=\alpha_{\min}+u_{t,1}(\alpha_{\max}-\alpha_{\min}),\\
\ell_t
&=\ell_{\min}+u_{t,2}(\ell_{\max}-\ell_{\min}),\\
b_t
&=u_{t,3}b_{\max},\\
s_t
&=\min\left\{1,\frac{0.98}{\alpha_t+\ell_t}\right\},\\
(\alpha_t,\ell_t)
&\leftarrow(s_t\alpha_t,s_t\ell_t).
\end{aligned}
\end{equation}
DQN uses the same decoder after selecting one point from a $5^3=125$ normalized action grid.

\paragraph{Unsafe supply, artificial liquidity, and market clearing.}
The implementation records how much invalid supply is economically backed and how much liquidity it creates:
\begin{equation}
\begin{aligned}
X_t
&=\max\Biggl\{\sum_i R^{rejected}_{i,t},\sum_i X^{excess}_{i,t},\\
&\qquad\quad
\max\Bigl(\sum_iP^{reported}_{i,t}-\sum_iP^{max}_{i,t},0\Bigr)\Biggr\},\\
\beta_t
&=\frac{\min(\alpha_t+\ell_t,0.98)}{0.98},\\
A^{unsafe}_t
&=\beta_tX_t,\\
G^{backed}_t
&=G^v_t+A^{unsafe}_t,\\
A^{artificial}_t
&=\ell_tA^{unsafe}_t.
\end{aligned}
\end{equation}
The executed market transition is then
\begin{equation}
\begin{aligned}
\widetilde{Q}_t
&=Q^d_t\max(0.75,1-0.80b_t),\\
L^{avail}_t
&=L_t+\ell_tG^{backed}_t,\\
M_t
&=\min(L^{avail}_t,\widetilde{Q}_t),\\
L_{t+1}
&=\max(L^{avail}_t-M_t,0),\\
\sigma^{exec}_t
&=\frac{\widetilde{Q}_t}{\max(L^{avail}_t+0.05,0.05)}.
\end{aligned}
\end{equation}

\paragraph{Episode, fairness, and audit metrics.}
The evaluation code aggregates step records into rollout-level metrics:
\begin{equation}
\begin{aligned}
R_e
&=\sum_{t\in e}R_t,\\
V_e
&=\frac{1}{24}\sum_{t\in e}V_t,\\
J_e
&=\frac{1}{23}\sum_{t=2}^{24}\|a_t-a_{t-1}\|_1,\\
\bar{\sigma}_e
&=\frac{1}{24}\sum_{t\in e}\sigma^{exec}_t,\\
A^{artificial}_e
&=\sum_{t\in e}A^{artificial}_t.
\end{aligned}
\end{equation}
City-level reward imbalance and agentic intervention strength are reported as
\begin{equation}
\begin{aligned}
F_e
&=\frac{\mathrm{Var}_c\left(\sum_{t\in e}r^{city}_{c,t}\right)}
{\max\left(\frac{1}{|C|}\sum_c\left|\sum_{t\in e}r^{city}_{c,t}\right|,10^{-9}\right)},\\
\delta_t
&=\|a^{final}_t-a^{original}_t\|_1,\\
\mathrm{audit\_rate}
&=\frac{N^{audit}}{24},\\
\mathrm{revision\_rate}
&=\frac{N^{revise}}{\max(N^{audit},1)}.
\end{aligned}
\end{equation}

\section{Data Provenance and Evaluation Matrix}
\label{app:raw-data-provenance}

Table~\ref{tab:app-raw-data-provenance} summarizes the benchmark input data used in this paper and documents the provenance of each raw file before policy evaluation. The primary dataset is a benchmark instance covering Beijing, Shanghai, Chengdu, Shenzhen, and Hangzhou from \texttt{2026-04-01} to \texttt{2026-05-01}. It contains 720 hourly timestamps and integrates cached city-level weather, node-level PV parameters, reported generation records, FDIA and verification labels, aggregate market-liquidity states, and generated P2P transaction traces. 

The evaluation artifacts reported in the paper are derived from these raw inputs through a fixed evaluation pipeline. The stored evidence records outcomes at multiple granularities, including episode-level metrics for rollout rewards and safety outcomes, hourly action rows for executed policy decisions, city-hour policy rows for spatial market states, violation, and fairness calculations, and LLM governance logs for Planner/Auditor decisions. Overall, the evaluation matrix contains 1,620 episode-level metric rows, 38,880 hourly action rows, 194,400 city-hour policy rows, and 12,960 LLM governance log rows.

\begin{table*}[h]
\centering
\caption{Raw benchmark data files and collection or generation methods. Row counts exclude CSV headers.}
\label{tab:app-raw-data-provenance}
\begin{tabular}{p{0.23\textwidth}r p{0.24\textwidth} p{0.34\textwidth}}
\toprule
File & Rows & Key fields & Collection or generation method \\
\midrule
\texttt{open\_meteo\_weather.json} & 5 city payloads & hourly time, temperature, shortwave radiation, latitude, longitude, timezone & Retrieved by \texttt{generate\_monthly\_datasets.py} from the Open-Meteo Historical Weather API for Beijing, Shanghai, Chengdu, Shenzhen, and Hangzhou; cached locally and reused on later runs. \\
\texttt{urban\_energy\_nodes.csv} & 50 & node id, city, latitude, longitude, panel area, efficiency, temperature coefficient, install date & Generated with seed 20260511: 10 nodes per city, coordinates jittered around city centers, panel areas sampled from 18--64 m$^2$, efficiencies from 0.176--0.226, temperature coefficients from -0.0046 to -0.0032, and install dates from 2020-01-01 to 2024-05-31. \\
\texttt{spatiotemporal\_generation.csv} & 36,000 & timestamp, node id, city, irradiance, air temperature, $P^{max}$, reported power, FDIA flag, verification status & Generated hourly for 50 nodes over 720 hours from cached weather, pvlib clear-sky and solar-position calculations, node-specific PV parameters, inverter derate, and noise. Exactly 1,800 rows are FDIA-injected and marked rejected. \\
\texttt{market\_liquidity.csv} & 720 & timestamp, total verified MW, SolarChain liquidity, baseline liquidity, SolarChain slippage, baseline slippage & Aggregated from verified generation. The implementation uses $L^{SC}=0.92G^v+0.018$, $L^{base}=0.61G^v+0.008$, $\sigma^{SC}=0.18/(L^{SC}+0.045)$, and $\sigma^{base}=0.31/(L^{base}+0.028)$. \\
\texttt{p2p\_trades.csv} & 1,185 & trade id, timestamp, factory id, city, energy purchased, tokens burned, exergy dissipated & Generated for daylight hours with positive verified supply. Each active hour can create up to three factory purchases; purchase size is the minimum of a SolarChain-liquidity share and a verified-generation share, with token burn and exergy fields sampled from configured multipliers. \\
\bottomrule
\end{tabular}
\end{table*}

\begin{table*}[h]
\centering
\small
\caption{Reproducibility protocol for the LLM Planner/Auditor layer.}
\label{tab:llm-governance-protocol}
\begin{tabular}{p{0.20\textwidth}p{0.74\textwidth}}
\toprule
Component & Specification \\
\midrule

LLM configuration &
Both Planner and Auditor use a fixed model, \texttt{ChatGPT 5.5 mini}, specified before evaluation and held constant across seeds, policies, episodes, and steps. The LLM is used only at evaluation time and is not involved in RL training. \\

Prompting &
Planner and Auditor use fixed system/user prompt templates released with the code. The templates are unchanged across runs; only the episode-level or step-level context is substituted into the prompt. \\

Planner output schema &
The Planner is called once per episode and must return
\texttt{governance\_mode}, \texttt{action\_bounds}, \texttt{audit\_policy}, and \texttt{rationale}. Bounds cover $\alpha_t$, $\ell_t$, and $b_t$; the audit policy specifies trigger thresholds, budget, target audit rate, and cooldown. \\

Auditor output schema &
The Auditor is called only for scheduler-eligible steps and must return
\texttt{decision}, \texttt{final\_action}, \texttt{risk\_assessment}, and \texttt{reason}. The decision is restricted to \texttt{approve} or \texttt{revise}. \\

Decoding and validation &
Structured-output parsing is used first; the chat JSON-schema fallback uses \texttt{temperature}=0. All outputs are parsed by strict schemas with extra fields forbidden. Valid numerical outputs are clipped and sanitized to benchmark constraints, including $\alpha_t+\ell_t\leq0.98$. \\

Failure handling &
Missing, unparsable, or schema-invalid LLM outputs fail the LLM evaluation run rather than being silently executed or replaced by free-form text. \\

Execution and logging &
Execution depends only on validated structured fields: \texttt{approve} keeps the bounded RL action, while \texttt{revise} executes the validated \texttt{final\_action}. Natural-language rationales are logged for audit evidence but do not control execution. \\

Determinism and baseline &
RL and environment seeds are fixed, but hosted LLM calls are not assumed to be bitwise deterministic. All accepted Planner/Auditor outputs are logged. A rule-based Planner/Auditor baseline uses the same scheduler, sanitizer, budget, cooldown, and logging interface. \\

\bottomrule
\end{tabular}
\end{table*}

\begin{table*}[h]
\centering
\caption{Detailed main-setting episode summaries. Reward, volume, violation, artificial liquidity, slippage, and action jitter are aggregated over 90 rollouts per policy.}
\label{tab:app-main-detail}
\begin{tabular}{lcccccccc}
\toprule
Policy & Rollouts & Reward Mean & Reward Min & Reward Max & Volume & Violation Mean & Violation Range & Artificial Liquidity \\
\midrule
DQN & 90 & -23.23 & -24.97 & -21.00 & 0.491 & 0.4688 & 0.4270--0.5048 & 0.1298 \\
Myopic & 90 & -23.08 & -24.60 & -21.39 & 0.456 & 0.4779 & 0.4434--0.5136 & 0.1009 \\
PPO & 90 & -22.35 & -24.20 & -20.14 & 0.470 & 0.4520 & 0.4060--0.4912 & 0.1453 \\
Random & 90 & -25.46 & -27.34 & -23.28 & 0.480 & 0.4771 & 0.4395--0.5147 & 0.1137 \\
SAC & 90 & -22.26 & -24.28 & -20.15 & 0.521 & 0.4461 & 0.4083--0.4879 & 0.0686 \\
Static & 90 & -23.38 & -24.86 & -21.65 & 0.514 & 0.4874 & 0.4527--0.5218 & 0.1803 \\
\bottomrule
\end{tabular}
\end{table*}

\begin{table*}[h]
\centering
\caption{Four-setting summaries for learned policies. Each row aggregates 90 rollouts across three seeds. These values support the paired physics-removal deltas and show how the LLM governance layer changes the setting-level risk profile.}
\label{tab:app-four-setting-learned}
\begin{tabular}{llrcccccc}
\toprule
Setting & Policy & Rollouts & Reward & Violation & Artificial Liquidity & Slippage & Jitter & Fairness \\
\midrule
RL & PPO & 90 & -22.35 & 0.4520 & 0.1453 & 0.0134 & 0.0607 & 0.0041 \\
RL & SAC & 90 & -22.26 & 0.4461 & 0.0686 & 0.0178 & 0.1435 & 0.0048 \\
RL & DQN & 90 & -23.23 & 0.4688 & 0.1298 & 0.0148 & 0.1118 & 0.0159 \\
RL+LLM & PPO & 90 & -22.44 & 0.4571 & 0.1379 & 0.0141 & 0.0829 & 0.0045 \\
RL+LLM & SAC & 90 & -22.32 & 0.4527 & 0.0759 & 0.0175 & 0.1145 & 0.0051 \\
RL+LLM & DQN & 90 & -23.51 & 0.4780 & 0.1406 & 0.0146 & 0.0939 & 0.0157 \\
RL no-physics & PPO & 90 & 0.27 & 0.4831 & 0.2238 & 0.0122 & 0.0135 & 0.0053 \\
RL no-physics & SAC & 90 & 0.26 & 0.4874 & 0.1925 & 0.0137 & 0.0432 & 0.0109 \\
RL no-physics & DQN & 90 & 0.32 & 0.4869 & 0.1874 & 0.0129 & 0.0259 & 0.0119 \\
RL+LLM no-physics & PPO & 90 & 0.22 & 0.4796 & 0.1977 & 0.0130 & 0.0366 & 0.0055 \\
RL+LLM no-physics & SAC & 90 & 0.30 & 0.4870 & 0.1908 & 0.0137 & 0.0362 & 0.0109 \\
RL+LLM no-physics & DQN & 90 & 0.27 & 0.4862 & 0.1820 & 0.0132 & 0.0342 & 0.0119 \\
\bottomrule
\end{tabular}
\end{table*}

\begin{table*}[h]
\centering
\caption{Aggregate learned-policy means by setting. Each row averages PPO, SAC, and DQN over the same 270 rollouts. The no-physics settings remain substantially higher in artificial liquidity even when LLM governance is active.}
\label{tab:app-four-setting-aggregate}
\begin{tabular}{lcccccc}
\toprule
Setting & Reward & Violation & Artificial Liquidity & Slippage & Jitter & Fairness \\
\midrule
RL & -22.61 & 0.4557 & 0.1146 & 0.0153 & 0.1053 & 0.0083 \\
RL+LLM & -22.76 & 0.4626 & 0.1181 & 0.0154 & 0.0971 & 0.0084 \\
RL no-physics & 0.28 & 0.4858 & 0.2012 & 0.0129 & 0.0275 & 0.0094 \\
RL+LLM no-physics & 0.26 & 0.4843 & 0.1901 & 0.0133 & 0.0357 & 0.0094 \\
\bottomrule
\end{tabular}
\end{table*}

\begin{table*}[h]
\centering
\caption{Detailed LLM governance operation rates. Each row aggregates 2,160 agentic log steps from three seeds. The LLM failure count is zero for every row and is therefore not shown.}
\label{tab:app-llm-ops}
\begin{tabular}{llcccccc}
\toprule
Setting & Policy & Steps & Audit Rate & Revision Given Audit & Modification Rate & Mean Action $\Delta$ & Not Triggered \\
\midrule
Main & DQN & 2160 & 0.372 & 0.958 & 0.518 & 0.084 & 0.488 \\
Main & PPO & 2160 & 0.344 & 0.961 & 0.987 & 0.164 & 0.550 \\
Main & SAC & 2160 & 0.342 & 0.966 & 0.524 & 0.059 & 0.509 \\
No-physics & DQN & 2160 & 0.297 & 0.892 & 0.490 & 0.042 & 0.683 \\
No-physics & PPO & 2160 & 0.273 & 0.832 & 0.991 & 0.111 & 0.691 \\
No-physics & SAC & 2160 & 0.274 & 0.876 & 0.294 & 0.021 & 0.698 \\
\bottomrule
\end{tabular}
\end{table*}

\begin{table*}[h]
\centering
\caption{Detailed LLM post-hoc trigger diagnostics and mean Planner thresholds. Trigger columns report the fraction of logged steps satisfying each trigger condition.}
\label{tab:app-llm-triggers}
\begin{tabular}{llcccccccc}
\toprule
Setting & Policy & Physics Trig. & Gap Trig. & Slippage Trig. & Jitter Trig. & Viol. Thr. & Gap Thr. & Jitter Thr. & Slip Thr. \\
\midrule
Main & DQN & 0.120 & 0.440 & 0.168 & 0.230 & 0.141 & -0.108 & 0.177 & 2.960 \\
Main & PPO & 0.121 & 0.440 & 0.171 & 0.171 & 0.142 & -0.112 & 0.174 & 2.961 \\
Main & SAC & 0.120 & 0.441 & 0.137 & 0.211 & 0.142 & -0.102 & 0.175 & 3.061 \\
No-physics & DQN & 0.116 & 0.441 & 0.170 & 0.049 & 0.143 & -0.099 & 0.170 & 3.028 \\
No-physics & PPO & 0.113 & 0.441 & 0.126 & 0.078 & 0.144 & -0.103 & 0.185 & 3.140 \\
No-physics & SAC & 0.105 & 0.441 & 0.145 & 0.048 & 0.146 & -0.102 & 0.181 & 3.111 \\
\bottomrule
\end{tabular}
\end{table*}

\begin{table*}[h]
\centering
\caption{Paired episode-level effects of adding the LLM Planner/Auditor layer. Deltas are computed as RL+LLM minus raw RL under matched seed, policy, and episode.}
\label{tab:app-agentic-paired-deltas}
\begin{tabular}{llccccc}
\toprule
Comparison & Policy & $\Delta$ Reward & $\Delta$ Action Jitter & $\Delta$ Spatial Fairness & $\Delta$ Artificial Liquidity & $\Delta$ Slippage \\
\midrule
Main & DQN & -0.277 $\pm$ 0.503 & -0.0179 $\pm$ 0.0274 & -0.0002 & 0.0109 & -0.0001 \\
Main & PPO & -0.094 $\pm$ 0.293 & 0.0223 $\pm$ 0.0154 & 0.0004 & -0.0074 & 0.0007 \\
Main & SAC & -0.062 $\pm$ 0.268 & -0.0290 $\pm$ 0.0198 & 0.0003 & 0.0073 & -0.0003 \\
No-physics & DQN & -0.045 $\pm$ 0.065 & 0.0083 $\pm$ 0.0110 & 0.0000 & -0.0054 & 0.0002 \\
No-physics & PPO & -0.043 $\pm$ 0.097 & 0.0231 $\pm$ 0.0166 & 0.0002 & -0.0261 & 0.0008 \\
No-physics & SAC & 0.038 $\pm$ 0.089 & -0.0070 $\pm$ 0.0154 & 0.0000 & -0.0017 & 0.0000 \\
\bottomrule
\end{tabular}
\end{table*}

\begin{table*}[h]
\centering
\caption{Step-level action distributions for learned policies. Each cell reports mean with 5th--95th percentile range in parentheses.}
\label{tab:app-action-distributions}
\begin{tabular}{llccc}
\toprule
Setting & Policy & Reward Ratio & Liquidity Ratio & Burn Rate \\
\midrule
Main & DQN & 0.320 (0.050--0.500) & 0.517 (0.200--0.762) & 0.081 (0.000--0.150) \\
Main & PPO & 0.050 (0.049--0.050) & 0.401 (0.200--0.931) & 0.104 (0.000--0.200) \\
Main & SAC & 0.169 (0.058--0.432) & 0.433 (0.212--0.738) & 0.008 (0.001--0.028) \\
No-physics & DQN & 0.216 (0.049--0.308) & 0.759 (0.672--0.931) & 0.080 (0.000--0.200) \\
No-physics & PPO & 0.049 (0.049--0.050) & 0.850 (0.200--0.931) & 0.045 (0.000--0.200) \\
No-physics & SAC & 0.215 (0.079--0.363) & 0.764 (0.617--0.898) & 0.011 (0.002--0.025) \\
LLM Main & DQN & 0.321 (0.050--0.500) & 0.575 (0.250--0.762) & 0.070 (0.000--0.150) \\
LLM Main & PPO & 0.065 (0.050--0.124) & 0.480 (0.250--0.900) & 0.071 (0.000--0.150) \\
LLM Main & SAC & 0.166 (0.060--0.402) & 0.470 (0.250--0.747) & 0.009 (0.001--0.027) \\
LLM No-physics & DQN & 0.217 (0.050--0.350) & 0.746 (0.500--0.900) & 0.069 (0.000--0.150) \\
LLM No-physics & PPO & 0.062 (0.050--0.100) & 0.799 (0.250--0.900) & 0.028 (0.000--0.150) \\
LLM No-physics & SAC & 0.212 (0.083--0.343) & 0.764 (0.634--0.888) & 0.011 (0.002--0.025) \\
\bottomrule
\end{tabular}
\end{table*}

\end{document}